\documentclass[journal]{IEEEtran}

\usepackage[table,xcdraw]{xcolor}
\usepackage{soul}
\definecolor{light_color}{rgb}{0.9,0.95,1}
\sethlcolor{light_color}
\usepackage{multirow}
\usepackage[colorlinks,linkcolor=blue,citecolor=blue]{hyperref}
\usepackage{cite}
\usepackage[pdftex]{graphicx}
\usepackage{array}
\usepackage[caption=false,font=normalsize,labelfont=sf,textfont=sf]{subfig}
\usepackage{svg}
\usepackage{amssymb}
\usepackage{xspace}
\usepackage{enumerate}
\usepackage[normalem]{ulem}
\useunder{\uline}{\ul}{}
\usepackage{pifont}
\usepackage{newtxtext}
\usepackage{algorithm}
\usepackage{algorithmic}
\usepackage[T1]{fontenc}
\usepackage{amsmath}
\usepackage{amsthm}

\usepackage[capitalize]{cleveref}
\crefname{section}{Sec.}{Secs.}
\crefname{table}{Tab.}{Tabs.}
\Crefname{section}{Section}{Sections}
\Crefname{table}{Table}{Tables}

\makeatletter
\DeclareRobustCommand\onedot{\futurelet\@let@token\@onedot}
\def\@onedot{\ifx\@let@token.\else.\null\fi\xspace}

\def\eg{\emph{e.g}\onedot} 
\def\ie{\emph{i.e}\onedot}

\def\wrt{w.r.t\onedot}

\makeatother

\def\ISPRS{\textbf{ISPRS}}
\def\DFC{\textbf{DFC2023}}
\def\DELIVER{\textbf{DELIVER}}
\def\wrtSOTA{\emph{\wrt SOTA}}

\def\MtL{M3L~\cite{maheshwari2024missing}}
\def\IMLT{IMLT~\cite{chen2024novel}}
\def\MAGIC{MAGIC~\cite{zheng2024centering}}
\def\MuSS{MuSS~\cite{wang2024multisenseseg}}

\newcommand{\GCOMMENT}[1]{
	\textcolor{gray}{\# #1}
}

\begin{document}
	
	\title{SGMA: Semantic-Guided Modality-Aware Segmentation for Remote Sensing with Incomplete Multimodal Data}
	
	\author{Lekang~Wen,
		Liang~Liao,~\IEEEmembership{Senior Member,~IEEE}, 
		Jing~Xiao,~\IEEEmembership{Senior Member,~IEEE},
		and~Mi~Wang,~\IEEEmembership{Senior Member,~IEEE}
		\thanks{L. Wen and M. Wang are with the State Key Laboratory of Information Engineering in Surveying, Mapping and Remote Sensing, Wuhan University, e-mail: wenlk3@whu.edu.cn, wangmi@whu.edu.cn.}
		\thanks{L. Liao is affiliated with the Hangzhou Institute of Technology, Xidian University, and can be reached at e-mail: liaoliang01@xidian.edu.cn.}
		\thanks{J. Xiao is with the School of Artificial Intelligence, Wuhan University, e-mail: jing@whu.edu.cn.}
		\thanks{Corresponding author: Jing Xiao.}
	}

	\markboth{Journal of \LaTeX\ Class Files,~Vol.~13, No.~9, September~2014}%
	{Shell \MakeLowercase{\textit{et al.}}: Bare Demo of IEEEtran.cls for Journals}
	
	\maketitle
	
	\begin{abstract}
		Multimodal semantic segmentation integrates complementary information from diverse sensors for remote sensing Earth observation. However, practical systems often encounter missing modalities due to sensor failures or incomplete coverage, termed Incomplete Multimodal Semantic Segmentation (IMSS). IMSS faces three key challenges: (1) \textit{multimodal imbalance}, where dominant modalities (\eg, RGB) suppress fragile ones (\eg, DSM, NIR, SAR); (2) \textit{intra-class variation} in scale, shape, and orientation across modalities; and (3) \textit{cross-modal heterogeneity} with conflicting cues producing inconsistent semantic responses. Existing methods rely on contrastive learning or joint optimization, which risk over-alignment, discarding modality-specific cues or imbalanced training, favoring robust modalities, while largely overlooking intra-class variation and cross-modal heterogeneity. To address these limitations, we propose the \textbf{Semantic-Guided Modality-Aware (SGMA)} framework, which ensures balanced multimodal learning while reducing intra-class variation and reconciling cross-modal inconsistencies through semantic guidance. SGMA introduces two complementary plug-and-play modules: (1) \textbf{Semantic-Guided Fusion (SGF)} module extracts multi-scale, class-wise semantic prototypes that capture consistent categorical representations across modalities, estimates per-modality robustness based on prototype-feature alignment, and performs adaptive fusion weighted by robustness scores to mitigate \textit{intra-class variation} and \textit{cross-modal heterogeneity}; (2) \textbf{Modality-Aware Sampling (MAS)} module leverages robustness estimations from SGF to dynamically reweight training samples, prioritizing challenging samples from fragile modalities to address \textit{modality imbalance}. Extensive experiments across multiple datasets and backbones demonstrate that SGMA consistently outperforms state-of-the-art methods, with particularly significant improvements in fragile modalities.
	\end{abstract}
	
	\begin{IEEEkeywords}
		Multimodal semantic segmentation, incomplete multimodality, remote sensing, multimodal learning
	\end{IEEEkeywords}

	\IEEEpeerreviewmaketitle
	
	\section{Introduction}

	\IEEEPARstart{T}{he} visual information captured by various sensors onboard remote sensing satellites is fundamental to a wide range of Earth observation tasks, including land use classification~\cite{zhang2024realtime,zhou2025advancing,ye2025tuple}, ecological monitoring~\cite{chen2026detracker,palharini2025challenges,nsr}, and urban planning~\cite{cetin2024determination,jadhav2024spatiotemporal,luojia}. 
	With the increasing availability of multimodal remote sensing data, such as optical imagery (RGB), near-infrared imagery (NIR), digital surface models (DSM), and synthetic aperture radar (SAR), Multimodal Semantic Segmentation (MSS) has become a key technique to integrate information from diverse sensors
	\cite{sti,albanwan2024image,vivone2025deep}
	for enhancing scene understanding under challenging conditions, such as cloud cover, shadows, or nighttime. For instance, in nighttime scenarios, optical imagery often suffers from degraded performance due to insufficient visual cues, while complementary modalities such as SAR or NIR can compensate for the inherent limitations of optical sensors, thereby improving the accuracy and reliability of remote sensing analysis.
	
	Despite notable progress in MSS, most existing approaches rely on fixed modality combinations, typically pairing RGB with another modality~\cite{PGDENet,MMSMCNet,zhang2025tagfusion}, which leads to significant performance degradation when any modality is absent. Such modality-missing situations frequently arise in remote sensing applications due to sensor malfunctions or incomplete regional coverage. To mitigate this problem, 
	Incomplete Multimodal Semantic Segmentation (IMSS) has emerged to enable robust performance under diverse and incomplete modality configurations. 
	Early studies in IMSS adopted modality dropout~\cite{ shi2024passion, maheshwari2024missing}, where some input modalities are randomly dropped during training. However, they fail to learn discriminative representations for fragile or less dominant modalities. Another line of work explored Masked Autoencoders (MAE)~\cite{he2022masked} and its variants~\cite{du2024multimodal,sosa2025multimae} by reconstructing masked inputs to enhance adaptability. However, these methods primarily focus on low-level appearance recovery rather than high-level semantic learning, and typically require finetuning for specific modality subsets, limiting their generalization to arbitrary modality-missing scenarios. More recent methods~\cite{chen2024novel,zheng2024learning,zheng2024centering} leverage contrastive learning to enforce cross-modal feature consistency. However, forcing robust modalities to align with weak ones may suppress modality-specific features and reduce discriminative capacity.
	
	\begin{figure*}[htb]
		\centering
		\includegraphics[width=1\linewidth]{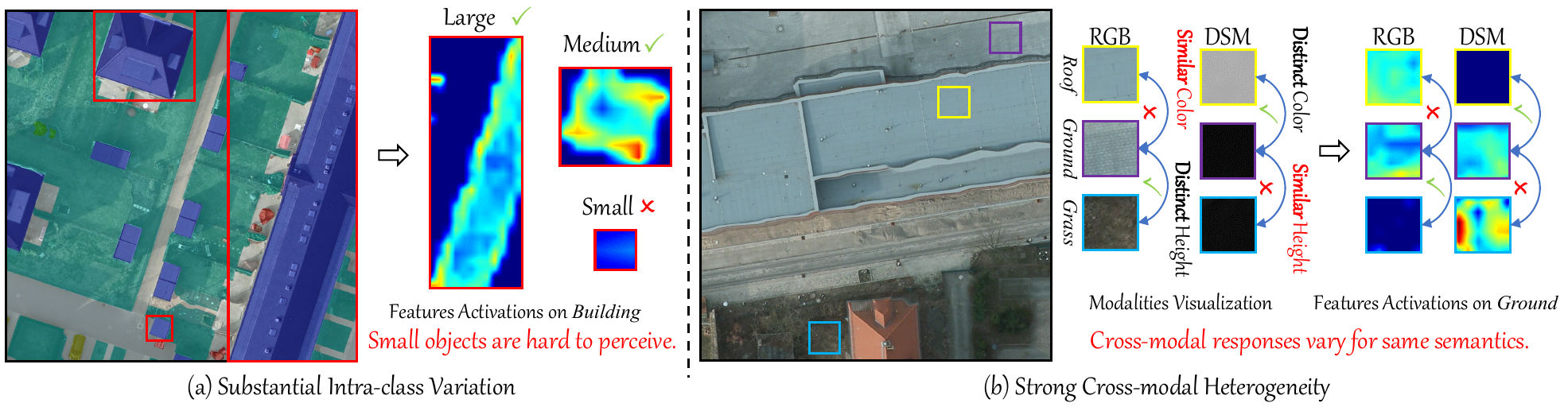}
		\caption{
			Illustration of the challenges of (a) substantial intra-class variation and (b) strong cross-modal heterogeneity in incomplete multimodal semantic segmentation for remote sensing. (a) The same semantic category, \textit{building}, exhibits large differences in scale and orientation, which complicates the extraction of consistent category features. (b) Different modalities produce conflicting cues: in RGB imagery, \textit{roof} and \textit{ground} share similar colors but differ in DSM height, while \textit{ground} and \textit{grass} have similar heights in DSM but distinct colors in RGB, making semantic correspondence challenging.
		}
		\label{fig: story}

	\end{figure*}

	In the remote sensing domain, IMSS is further complicated by two intrinsic challenges. First, objects within the same category often exhibit substantial intra-class variation in scale, orientation, and shape. 
	As illustrated in \cref{fig: story}(a), \textit{building} instances within a large-scale scene can appear in vastly different sizes and orientations. Notably, smaller buildings exhibit weak and sparse feature activations compared to their larger counterparts, which reveals the fundamental challenge of establishing category-consistent representations across diverse object scales.
	To address this issue, existing approaches empoly dynamic convolutions~\cite{wang2024multimodal,he2025semibacon,feng2025df2rq}, long-range attention mechanisms~\cite{SDM-Car,feng2025mdfnet,khan2025coastal,zhang2025establishing}, and multi-scale feature fusion~\cite{MFFENet,ma2024adjacentscale} to enlarge receptive fields, but their applicability to multimodal data is limited by fixed channel allocations, high computational costs, and the complexity of cross-scale concatenation, which restrict scalability to arbitrary modality inputs.
	Second, different modalities exhibit strong heterogeneity in inter-class responses. 
	As shown in \cref{fig: story}(b), regions with similar appearances extend to produce indistinguishable features within individual modalities. However, this limitation can be compensated for through modality heterogeneity: visually similar regions, such as \textit{roof} and \textit{ground}, are easily distinguished in DSM by elevation, while \textit{grass} and \textit{ground} with similar heights are separable in RGB by color. This heterogeneity makes establishing reliable semantic correspondences across modalities highly non-trivial.
	Early cross-modal fusion methods, such as summation~\cite{audebert2018rgb, hong2021more} and concatenation~\cite{hong2023crosscity, liu2023mmglots}, provide intuitive strategies but apply fixed-weight fusion, causing dominant modalities to overshadow information from weaker ones. Self-attention-based methods~\cite{fan2024elevation,chen2024novel} enhance cross-modal feature interactions but fail to quantify modality-specific contributions for weighted fusion, and their computational complexity scales quadratically with the number of modalities. Hand-crafted weighting schemes~\cite{yu20243d, lan2024uav, gao2025mmradnet, sotomayor2025mapping} based on modality-specific metrics can achieve effective fusion for particular combinations, such as elevation guidance for building recognition~\cite{yu20243d,gao2025mmradnet} or infrared band mapping for vegetation classification~\cite{lan2024uav, sotomayor2025mapping}, yet such biases limit their generalization across diverse modalities.
	Overall, these limitations motivate the development of a unified IMSS framework that not only mitigates modality imbalance and scales to arbitrary multimodal inputs but also explicitly addresses the dual challenges of intra-class variation and cross-modal heterogeneity in remote sensing imagery.
	
	In this work, we propose the \textbf{S}emantic-\textbf{G}uided \textbf{M}odality-\textbf{A}ware (SGMA) segmentation framework for IMSS, which ensures balanced contributions of both robust and fragile modalities while reducing intra-class variation and reconciling inconsistent cross-modal responses. 
	The proposed framework is built upon three key ideas. First, it dynamically evaluates modality robustness and adaptively prioritizes learning from fragile modalities to prevent robust modalities from dominating the learning process, thereby alleviating multimodal imbalance. Second, it compresses multimodal features into global semantic prototypes, which provide comprehensive receptive fields and enhance category consistency, thereby reducing intra-class variation. Third, it introduces semantic-guided attention fusion, where semantic prototypes serve as queries to adaptively aggregate features from arbitrary modalities, ensuring that each modality contributes optimally according to its distinctive characteristics.
	
	To achieve this, SGMA designs two plug-and-play modules: (1) Semantic-Guided Fusion (SGF): SGF extracts global semantic prototypes as intermediate anchors, directly linking pixel representations to their semantic centroids. This design reduces intra-class variance by strengthening category consistency. These prototypes further act as queries in a semantic-guided attention fusion mechanism, where modality-specific features serve as key–value pairs. Through attention-weighted aggregation, each modality contributes most effectively to the categories it best characterizes, while the learned attention weights provide an explicit measure of modality robustness. (2) Modality-aware Sampling (MAS): MAS leverages robustness scores from SGF to dynamically adjust training sample probabilities across modalities. By increasing the learning frequency of fragile modalities, MAS alleviates multimodal imbalance, strengthens feature learning, and enhances adaptability under arbitrary modality-missing scenarios.
	
	In summary, our main contributions are threefold:
	
	\begin{itemize}
		\item We propose the Semantic-Guided Modality-Aware (SGMA) framework to address three critical issues in IMSS for remote sensing: modality imbalance, intra-class variation, and cross-modal heterogeneity, thereby enhancing the robustness and generalization of semantic segmentation under incomplete modality conditions.
		
		\item We introduce the Semantic-Guided Fusion (SGF) module, which leverages global semantic prototypes as intermediate anchors to establish robust cross-modal correspondences, enabling effective feature aggregation across arbitrary modality combinations while providing explicit quantification of robustness.
		
		\item We design the Modality-Aware Sampling (MAS) module, which adaptively rebalances multimodal learning based on robustness assessment, significantly improving the representation quality of fragile modalities without requiring modality-specific architectural modifications.
		
	\end{itemize}

    The rest of this paper is organized as follows. \cref{sec: related_work} reviews related work on multimodal learning and incomplete multimodal learning, identifying limitations of existing approaches. \cref{sec: methodology} formally defines the IMSS problem and presents the SGMA framework with its two core modules: SGF for semantic-guided fusion and MAS for modality-aware sampling. \cref{sec: experiments} validates our method across three datasets through qualitative and quantitative comparisons and ablation studies. \cref{sec: conclusion} summarizes our contributions and discusses future directions.
	
	\section{Related Work}\label{sec: related_work}

    In this section, we briefly review related work in multimodal learning and incomplete multimodal learning, identifying limitations of existing approaches in addressing modality imbalance, intra-class variation, and cross-modal heterogeneity.

	\subsection{Multimodal Learning}
	
	Multimodal learning aims to enhance the accuracy and robustness of scene understanding by integrating information from different sensors. Early studies focused on fusing RGB with complementary modalities, such as SAR~\cite{xiao2025multimodal,xu2025semantic}, DSM~\cite{ma2024multilevel,yu20243d, feng2025ftransdeeplab}, and NIR~\cite{gao2024nwpumoc, lan2024uav, sotomayor2025mapping}.  
	For instance, NWPU-MOC~\cite{gao2024nwpumoc} employs a dual attention mechanism to fuse RGB and NIR features for fine-grained aerial image analysis, where NIR provides distinctive spectral characteristics for vegetation and water bodies while RGB captures spatial structures.
	With the advancement of sensor technologies, dual-modal learning has expanded to multimodal learning, where approaches can be broadly categorized into symmetric branch distribution~\cite{reza2024mmsformer,wang2024multisenseseg,zheng2024learning} and asymmetric branch distribution~\cite{zhang2023delivering,gao2024global,zhang2025tagfusion} architectures. The former treats all modalities equally. 
	The latter selects the most robust modality as the primary modality and incorporates others as auxiliary inputs. 
	 Beyond sensor-based modalities, language is the pivotal semantic modality, driving the rapid development of vision-language foundation models~\cite{kuckreja2024geochat, hu2025rsgpt}. Leveraging the Transformer architecture for multimodal fusion, these models enable complex vision-language understanding~\cite{zhang2024multilevel, li2025mffnet} and cross-modal reasoning~\cite{ding2025modalinvariant, zhang2025tfdet}, extending their applications to diverse remote sensing tasks such as change detection~\cite{zan2025openvocabulary,gao2025combining} and visual question answering~\cite{wang2024earthvqa,wang2024rsadapter}.
	
	These methods demonstrate impressive performance when all modalities are available. However, most of them are generally designed for a fixed modality combination. As a result, they will degenerate or fail in incomplete multimodal scenarios, where arbitrary modalities may be missing or corrupted.

	\subsection{Incomplete Multimodal Learning}
	
	Incomplete multimodal learning aims to maintain reliable performance when certain modalities are missing,
	with extensive research spanning medical imaging~\cite{wang2023learnable,shi2024passion}, remote sensing~\cite{du2024multimodal,chen2024novel,sosa2025multimae}, and autonomous driving~\cite{maheshwari2024missing,zheng2024centering,wen2025charm}.
	Early research addressed this challenge using modality dropout~\cite{shi2024passion, maheshwari2024missing}, which randomly removes input sources during training to enhance robustness. \MtL{} extends this idea by leveraging stochastic modality dropping to encourage modality-specific feature learning and mitigate information loss when inputs are missing. Another direction has investigated Masked Autoencoders (MAE)~\cite{he2022masked} and its variants~\cite{du2024multimodal,sosa2025multimae}, which pretrain models to reconstruct masked inputs for better adaptation to multimodal settings. 
	More recently, contrastive learning has been adopted to explicitly align cross-modal representations~\cite{wang2023learnable,chen2024novel,zheng2024centering}. For example, \IMLT{} integrates contrastive learning with masked pretraining to enhance cross-modal consistency and improve generalization, making the first IMSS-oriented method in remote sensing. \MAGIC{} further advances this paradigm by grouping modalities into robust and fragile sets, optimizing them jointly, and aligning their features with a cosine-based loss to yield a representation space robust to arbitrary-modal configurations.
	
	Despite the notable progress, existing methods face critical limitations. Dropout-based~\cite{shi2024passion,maheshwari2024missing} and grouping strategies~\cite{zheng2024centering} increase missing-modality exposure but inadequately train fragile modalities. MAE-based approaches~\cite{du2024multimodal,sosa2025multimae} emphasize low-level reconstruction over discriminative features, while alignment methods~\cite{chen2024novel,zheng2024centering} may weaken modality-specific characteristics by enforcing uniform consistency. In contrast, SGMA formulates cross-modal interaction as class-level semantic attention, enabling interpretable and adaptive modality weighting. Global semantic prototypes anchor pixels to category centroids to reduce intra-class variance, while robustness-guided sampling dynamically prioritizes fragile modalities, ensuring balanced learning without architectural constraints.

	\section{Methodology}\label{sec: methodology}

    In this section, we describe the proposed SGMA framework for IMSS. We first formalize the problem setting and provide an overview of our framework. Next, we present the Semantic-Guided Fusion (SGF) and Modality-Aware Sampling (MAS) modules and their associated objective functions.

	\begin{figure*}[htb]
		\centering
		\includegraphics[width=1\linewidth]{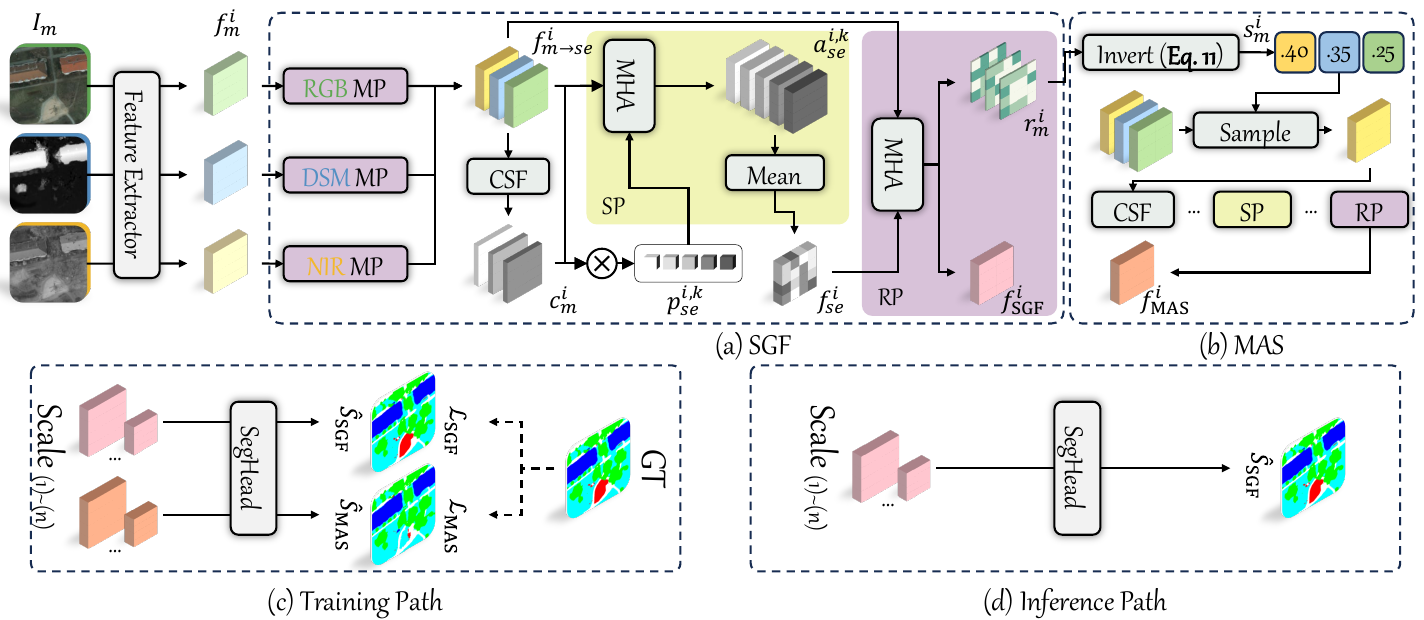}
		\caption{Overall framework of the proposed SGMA, which includes the Semantic-Guided Fusion (SGF) and Modality-Aware Sampling (MAS) modules.}
		\label{fig: framework}
	\end{figure*}
	
	\subsection{Problem Formulation of IMSS}

	We define a modality set $\mathcal{M}$ with $M$ elements, where the multimodal imagery input is denoted as $\mathcal{I} = \{I_m\}_{m \in \mathcal{M}}$, and the corresponding segmentation labels are denoted as $S$. During training, all modalities are assumed to be accessible; however, at inference, only a subset $\mathcal{I}_{\text{sub}} \subseteq \mathcal{I}$ may be available due to sensor failure or incomplete coverage. The objective of IMSS is to train a segmentation model that remains robust and effective across arbitrary modality combinations, even in the presence of missing inputs. 
	
	Models trained solely on complete multimodal supervision typically experience a significant drop in accuracy when inputs are absent. To address this, existing methods commonly adopt two strategies: (1) modality dropout, which stochastically samples a non-empty subset $\mathcal{I}_{\text{sub}} \subseteq \mathcal{I}$ from all possible modality combinations during each training iteration to simulate diverse incomplete-modality scenarios, and (2) contrastive alignment, which employs contrastive learning to encourage feature consistency across representations derived from different modality subsets. Formally, the model $\Phi$ is optimized with the following objective:
	\begin{equation}
		\begin{aligned}
			\min_{\Phi} \mathcal{L}_{\text{IMSS}} =\ 
			&\lambda_\text{seg}\,\mathcal{L}_{\text{seg}}(\Phi(\mathcal{I}_{\text{sub}}), S) \\
			+&\lambda_\text{con}\!\!\sum_{(m,m') \in \mathcal{P}} \mathcal{L}_{\text{con}}(f_m, f_{m'})),
		\end{aligned}
	\end{equation}
	where $\mathcal{P}=\{(m,m') \mid m,m'\in\mathcal{M}, m \neq m'\}$ is the set of modality pairs, $f_m$ and $f_{m'}$ are features of modalities $m$ and $m'$, $\mathcal{L}_{\text{seg}}$ is the segmentation loss, $\mathcal{L}_{\text{con}}$ enforces cross-modal alignment,  $\lambda_\text{seg}$ and $\lambda_\text{con}$ are trade-off weights.

	\subsection{Framework Overview}
	
	
	
	The overall framework of the proposed SGMA is illustrated in \cref{fig: framework}. We establish IMSS as a dual-branch optimization problem. As shown in \cref{fig: framework}(a), the Semantic-Guided Fusion (SGF) module addresses cross-modal semantic inconsistency through class-level context modeling. Specifically, SGF constructs global semantic prototypes for each class and employs the Spatial Perceptron (SP) to associate each pixel representation with its corresponding semantic centroid, ensuring semantic consistency across heterogeneous modalities and reducing intra-class variation. The Robustness Perceptron (RP) further addresses cross-modal heterogeneity through adaptive modality weighting based on reliability scores.
	As shown in Fig.~\ref{fig: framework}(b), the Modality-Aware Sampling (MAS) module leverages robustness scores from SGF to dynamically adjust sampling probabilities. By increasing the training frequency of fragile modalities, MAS alleviates multimodal imbalance and improves their representation quality.

	Taking three modalities of RGB (R), DSM (D), and NIR (N) modalities as an example, SGMA packs all input images $I_{m}$ into a mini-batch for efficient parallel computation. A shared-weight encoder $F$ extracts features for each modality independently across each scale:  
	\begin{equation}
		\{f_{m}^{i}\}_{i=1}^4=F(I_{m}).
	\end{equation}
	where $i$ denotes the scale index and the encoder $F$ produces features at 4 different scales.
	
	During training, as shown in \cref{fig: framework}(c), the feature $f_{m}^{i}$ from each modality is processed through both pathways to generate a semantic-guided fused feature $f_{\text{SGF}}^i$ and a modality-aware sampled feature $f_{\text{MAS}}^i$. These features are then fed into a segmentation head to produce the segmentation prediction $\hat{S}_{\text{SGF}}$ from SGF and the prediction $\hat{S}_{\text{MAS}}$ from MAS for joint optimization. In the inference stage, as shown in \cref{fig: framework}(d), only the SGF is retained, and the prediction $\hat{S}_{\text{SGF}}$ is the output segmentation map.

	\subsection{Semantic-Guided Fusion Module \label{sec: SGF}}
	
	To reduce intra-class variance and mitigate the impact of cross-modal heterogeneity, SGF computes global class-wise semantic representations for each modality, thereby suppressing irrelevant information. These representations are then used to query semantic class features at every scale, enabling an assessment of modality robustness and enhancing the representations of semantic features.
	
	\begin{figure}[htb]
		\centering
		\includegraphics[width=1\linewidth]{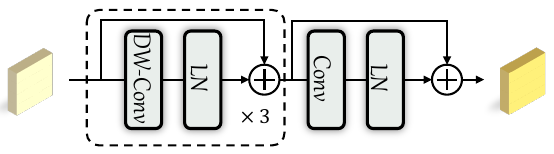}
		\caption{Structure of the Modality-specific Projector (MP).}
		\label{fig: mp}
	\end{figure}
	

	As illustrated in \cref{fig: framework}(a), SGF consists of four main components: 1) Modality-specific Projector (MP), which transforms modality-specific features into a unified semantic space through three depth-wise convolutions with kernel sizes of $11 \times 11$, $7 \times 7$, and $3 \times 3$, followed by a $1 \times 1$ point-wise convolution for semantic projection, as described in \cref{fig: mp}; 2) Class-aware Semantic Filter (CSF), which is essentially a shared $1 \times 1$ convolution that compresses modality-specific features from $C_i$ channels to classes $K$ channels, extracting class-level semantic representations; 3) Spatial Perceptron (SP), which leverages global semantic prototypes to query pixel-level features through multi-head attention (MHA), reducing intra-class variation via semantic-guided attention; and 4) Robustness Perceptron (RP), which evaluates modality reliability and performs adaptive fusion using multi-head attention to address cross-modal heterogeneity,
    sharing conceptual similarities with uncertainty-driven active learning~\cite{hertel2023probabilistic,wang2025semisupervised,valiuddin2025review}.
	
	Specifically, each modality feature $f_m^i$ is first transformed by its corresponding MP into the modality semantic feature $f_{m\rightarrow{se}}^i\in\mathbb{R}^{H_i\times W_i\times C_i}$, where $H_i$, $W_i$, and $C_i$ denote the height, width, and channel number of features at scale $i$. 
	Next, these features are filtered by the CSF that reduces channel dimensionality from $C_i$ to $K$ to compress high-level semantic information into class-specific representations. It produces the modality compact feature $c^i_m \in \mathbb{R}^{H_i \times W_i \times K}$, which is then used to obtain global semantic prototypes $p^{i,k}_{se} \in \mathbb{R}^C$ as follows:
	
	\begin{equation}
		\{p_{se}^{i,k}\}_{k=1}^{K} = \big[\{c_m^i\}_{m\in \mathcal{M}}\big] \otimes \big[\{f_{m\rightarrow{se}}^i\}_{m\in \mathcal{M}}\big]^T, \label{equ: prototype_construction}
	\end{equation}
	where $ \big[\cdot\big] $ denotes the concatenation of tensors in the set and $\otimes$ represents matrix multiplication. The concatenated tensor on the left, $ \big[\{c_m^i\}_{m\in \mathcal{M}}\big] $, has a shape of $ \mathbb{R}^{(H_i \times W_i \times M) \times K} $, while the concatenated tensor on the right, $ \big[\{f_{m\rightarrow{se}}^i\}_{m\in \mathcal{M}}\big] $, is of shape $ \mathbb{R}^{(H_i \times W_i \times M) \times C_i} $. This transformation converts dense multimodal features into global semantic prototypes, allowing high-level semantics to be aggregated around each category. In this way, semantic information becomes more concentrated, which helps to improve the discriminative power of the resulting representations.
	
	The prototypes $p_{se}^i$ are employed to query each local pixel location in the multimodal features through the SP:
	\begin{equation}
		a_{se}^{i,k} = \operatorname{MHA_{SP}}(q_i,k_i,v_i),
	\end{equation}
	where the query input is defined as
	\begin{equation}
		q_i = \operatorname{repeat}(p_{se}^i) \in \mathbb{R}^{(H_i \times W_i) \times K \times C},
	\end{equation}
	representing repeated global class prototypes broadcast to every spatial location. The key and value inputs are given by 
	\begin{equation} 
		\begin{gathered}
			k_i = v_i = \\
			\operatorname{rearrange}\left(\left[\{f_{m\rightarrow{se}}^i\}_{m\in \mathcal{M}}\right]\right) \in \mathbb{R}^{(H_i \times W_i) \times M \times C},
		\end{gathered}
	\end{equation}
	which correspond to the rearranged multimodal semantic features. Through the MHA layer, we obtain spatial semantic activation $ a_{se}^{i,k} \in \mathbb{R}^{H_i \times W_i \times C} $. $(H_i \times W_i)$ represents the flattened spatial dimensions to enable pixel-wise attention computation. The semantic-guided feature $f_{se}^i$ is computed by average across all $ K $ classes: 
	
	\begin{equation}
		f_{se}^i = \frac{1}{K} \sum_{k=1}^K a_{se}^{i,k}.
	\end{equation}
	
	Subsequently, $f_{se}^i$ serves as the foundation for modality robustness assessment through the RP. It evaluates the reliability and contribution of each modality by computing attention weights that reflect their robustness for different semantic classes. Specifically, the RP employs another MHA layer to simultaneously obtain the semantic-guided fused feature $f_\text{SGF}^i$ and attention weight maps to serve as modality robustness maps $\{r_m^{i}\}_{m \in \mathcal{M}}$:
	\begin{equation}
		f_\text{SGF}^i, \{r_m^{i}\}_{m \in \mathcal{M}} = \operatorname{MHA_{RP}}(q'_i, k'_i, v'_i),
	\end{equation}
	where the query input is defined as
	\begin{equation} 
		q'_i = \operatorname{repeat}(f_{se}^i) \in \mathbb{R}^{(H_i \times W_i) \times 1 \times C}.
	\end{equation}
	The key and value inputs are given by 
	\begin{equation} 
		k'_i = v'_i = \operatorname{rearrange}\left(\left[\{f_{m\rightarrow{se}}^i\}_{m\in \mathcal{M}}\right]\right) \in \mathbb{R}^{(H_i \times W_i) \times M \times C}.
	\end{equation} 
	RP uses semantic-guided features $f^i_{se}$ as queries. This design enables attention to measure \textit{how well each modality aligns with semantic class representations}, facilitating category-dependent reliability assessment. This differs from standard attention that only uses raw features as queries to measure feature-level similarity. The robustness scores $\{r_m^{i}\}_{m \in \mathcal{M}}$ quantify modality reliability across class and scale dimensions. For the class, by querying with semantic prototypes, \textit{these scores inherently capture class-wise contributions}, where DSM features yield high scores for structural classes while NIR excels for vegetation. For the scale, \textit{computing scores independently at each scale $i$ captures hierarchical contributions from fine-grained textures to coarse structures}. This differs from standard attention, which measures feature similarity without considering semantic or hierarchical context.
	
	Intuitively, higher attention weights reflect richer and more stable semantic content. Consequently, the maps $r_m^{i} \in \mathbb{R}^{H_i \times W_i}$ indicate the spatial reliability of each modality. This adaptive weighting ensures that modalities contribute most effectively based on their robustness, thereby alleviating cross-modal heterogeneity and improving the learned representation. 
	
	\subsection{Modality-aware Sampling Module \label{sec: MAS}}
	
	In multimodal fusion, imbalanced modality reliability often causes robust modalities to dominate, thereby suppressing the learning of fragile ones. To address this, MAS decouples the training of fragile modalities by sampling them more frequently. As shown in \cref{fig: framework}(b), MAS is designed to selectively sample less reliable modalities and exploit their potential through independent training. 
	
	Intuitively, a higher robustness score $ r_m^i $ indicates that the semantic information in modality $ m $ is easier to learn and distinguish, and thus requires less sampling emphasis. To formalize this intuition, we first apply pixel-wise conversion to all robustness maps $\{r_m^i\}_{m\in\mathcal{M}}$ to obtain inverted robustness maps $\{\hat{r}_m^i\}_{m\in\mathcal{M}}$:
	\begin{equation}
		\hat{r}_m^i = \frac{1/r_{m}^i}{\sum_{m\in\mathcal{M}}(1/r_m^i)}.
		\label{equ: softmax2softmin}
	\end{equation}
	
	The inverted robustness maps $\{\hat{r}_m^i\}_{m\in\mathcal{M}}$ are then spatially converted into scalars through average pooling to obtain modality sampling probabilities $\{s_m^i\}_{m\in\mathcal{M}}$.
	Since the robustness maps are derived from attention weight matrices that have already undergone \textit{SoftMax} normalization, the resulting values are relatively close to each other. Directly applying \textit{SoftMin} to these normalized values would further smooth the differences, causing sampling probabilities to converge toward a uniform distribution and undermining the selective sampling strategy. Instead, \cref{equ: softmax2softmin} is equivalent to \textit{SoftMin} operation on the original pre-\textit{SoftMax} values, offering computational efficiency by eliminating the need for original values, mathematical symmetry between \textit{SoftMax} and \textit{SoftMin} transformations, and algorithmic simplicity through direct reciprocal conversion. 

	During each training iteration, one modality feature at the current scale is stochastically selected based on these probabilities, where $s_{m}^i$ denotes the sampling probability for the semantic feature $f_{m \rightarrow se}^i$. The selected semantic feature follows the same path through the SGF and serves as the MAS output $f_\text{MAS}^i$. Fragile modalities with lower robustness receive higher sampling probabilities, promoting learning from challenging modalities to enhance their robustness.
	
	\begin{algorithm}[htb]
		\caption{SGMA Unified Training and Inference Procedure}
		\label{alg: frame}
		
	\begin{algorithmic}[1]
		\REQUIRE Multimodal images $\mathcal{I}=\{I_m\}_{m \in \mathcal{M}}$,ground truth $S$
		\ENSURE Segmentation $\hat{S}$ (inference) / Updated $\Theta$ (training)
		
		\STATE $\{f^i_m\}_{m\in\mathcal{M},i\in\{1,2,3,4\}} \gets \{F(I_m) | \forall m \in \mathcal{M}\}$
		
		\FOR{each scale $i$}
		\STATE \GCOMMENT{$(M,H_i,W_i,C_i)$}
		\STATE $\{f^i_{m \to se}\}_{m\in\mathcal{M}} \gets \{\text{MP}(f^i_m) | \forall m\}$
		
		\STATE \GCOMMENT{\textbf{Semantic-Guided Fusion (SGF)}}
		\STATE \GCOMMENT{$(M,H_i,W_i,K)$}
		\STATE $c^i \gets \text{CSF}(\{f^i_{m \to se}\}_{m\in\mathcal{M}})$
		\STATE \GCOMMENT{$(K,C_i)$}
		\STATE $p^i_{se} \gets \text{einsum}(c^i,\{f^i_{m \to se}\}_{m\in\mathcal{M}},\text{`mhwk,mhwc$\to$kc'})$
		\STATE \GCOMMENT{$(K,H_i,W_i,C_i)$}
		\STATE $a^i_{se} \gets \text{MHA}_{\text{SP}}(p^i_{se},\{f^i_{m \to se}\}_{m\in\mathcal{M}})$
		\STATE \GCOMMENT{$(H_i,W_i,C_i)$, $(M,H_i,W_i)$}
		\STATE $f^i_{\text{SGF}}, r^i \gets \text{MHA}_{\text{RP}}(a^i_{se},\{f^i_{m \to se}\}_{m\in\mathcal{M}})$
		
		\IF{training}
		\STATE \GCOMMENT{\textbf{Modality-Aware Sampling (MAS)}}
		\STATE $\hat{r}^i \gets \text{inv-softmax}(r^i,\text{dim=0})$
		\STATE \GCOMMENT{$(M,)$}
		\STATE $s^i_{prob} \gets \text{reduce}(\hat{r}^i,\text{`mhw$\to$m'},\text{`mean'})$
		\STATE $m^* \sim \text{sample}(\mathcal{M},s^i_{prob})$
		\STATE \GCOMMENT{$(H_i,W_i,C_i)$}
		\STATE $f^i_{\text{MAS}} \gets \text{SGF}(f^i_{m^* \to se})$
		\ENDIF
		\ENDFOR
		
		\IF{training}
		\STATE $\hat{S}_\text{SGF},\hat{S}_\text{MAS} \gets \text{SegHead}(\{f^i_{\text{SGF}},f^i_{\text{MAS}}\}_{i\in\{1,2,3,4\}})$
		\STATE $\mathcal{L} \gets \mathcal{L}_{\text{SGF}}(\hat{S}_\text{SGF},S) + \mathcal{L}_{\text{MAS}}(\hat{S}_\text{MAS},S)$
		\STATE Update $\Theta$ via backpropagation
		\ELSE
		\STATE $\hat{S} \gets \text{SegHead}(\{f^i_\text{SGF}\}_{i\in\{1,2,3,4\}})$
		\RETURN $\hat{S}$
		\ENDIF
	\end{algorithmic}
	\end{algorithm}
	
	To provide a comprehensive understanding of our framework, we present the complete training and inference procedure in Algorithm~\ref{alg: frame}. It explicitly delineates the distinct roles of SGF and MAS during different phases. Both modules are employed during training to enable robust multimodal fusion and fragile modality augmentation, while only SGF is activated during inference.
	
	\subsection{Objective Functions}
	
	Since $f_\text{SGF}^i$ and $f_\text{MAS}^i$ have the same shape, they can be packed into a batch at the same scale to accelerate parallel training. Output features from the SGF and MAS modules are packaged as input to SegHead. The loss function $\mathcal{L}_\text{SGF}$ for SGF and the loss function $\mathcal{L}_\text{MAS}$ for MAS both use cross-entropy loss for supervised training: 
	\begin{equation}
		\mathcal{L}_\text{SGF}=-\sum_{p\in S} S(p) \log{(\hat{S}_{\text{SGF}}(p))},
	\end{equation}
	\begin{equation}
		\mathcal{L}_\text{MAS}=-\sum_{p\in S} S(p) \log{(\hat{S}_{\text{MAS}}(p))},
	\end{equation}
	where $p$ is the pixel index for the segmentation map $S$. The loss function used for training the SGMA is formalized as follows:
	\begin{equation}
		\mathcal{L}_\text{IMSS} = \lambda_{\text{SGF}}\mathcal{L}_{\text{SGF}} + \lambda_{\text{MAS}}\mathcal{L}_{\text{MAS}},
	\end{equation}
	where $\lambda_{\text{SGF}}$ and $\lambda_{\text{MAS}}$ are balancing coefficients.
	
	\section{Experiments}\label{sec: experiments}

    In this section, we conduct comprehensive experiments to validate SGMA. We first describe the experimental setup, then present qualitative and quantitative comparisons across multiple datasets and backbones, and finally perform extensive ablation studies to analyze each component.

	\subsection{Experimental Setup}
	
	\subsubsection{Datasets}
	
    We evaluate our method on two multimodal remote sensing datasets: ISPRS Potsdam~\cite{rottensteiner2012isprs}, referred to as \ISPRS{}, and Data Fusion Contest 2023~\cite{persello20232023}, referred to as \DFC{}, and one autonomous driving dataset~\cite{zhang2023delivering}, referred to as \DELIVER{}. \ISPRS{} provides RGB, DSM, and NIR imagery with annotations for five categories (impervious surface, building, low vegetation, tree, and vehicle). The dataset contains 38 images split into 24 training and 14 validation samples. \DFC{} includes RGB, DSM, and SAR data with building annotations. It comprises 1773 images divided into 1419 training and 354 validation samples. \DELIVER{} offers RGB, Depth, Event, and LiDAR data with 25 semantic classes across diverse environmental conditions (sunny, cloudy, foggy, rainy, nighttime) and sensor failure scenarios. The dataset contains 3983 training and 2005 validation samples.
	
	\subsubsection{Baseline Methods}
	
	We compare our method with four advanced methods, \ie, (1) \MuSS{}, which concatenates all multimodal features and fuses them through linear layers for universal segmentation for MSS; (2) \MtL{}, which employs random modality dropping to train learnable parameters that retain modality-specific representations for IMSS; (3) \IMLT{}, which integrates contrastive learning to strengthen modality alignment and employs masked pretraining to enhance generalization for IMSS; (4) \MAGIC{}, which incorporates training with the most robust and fragile modalities to improve segmentation framework resilience for IMSS. For equality, the backbones of all methods are set to PVT-v2-b2~\cite{wang2022pvt} and ResNet-50~\cite{he2016deep}. All methods were retrained from scratch under identical experimental settings.
	
	\begin{figure*}[htb]
		\centering
		\includegraphics[width=1\linewidth]{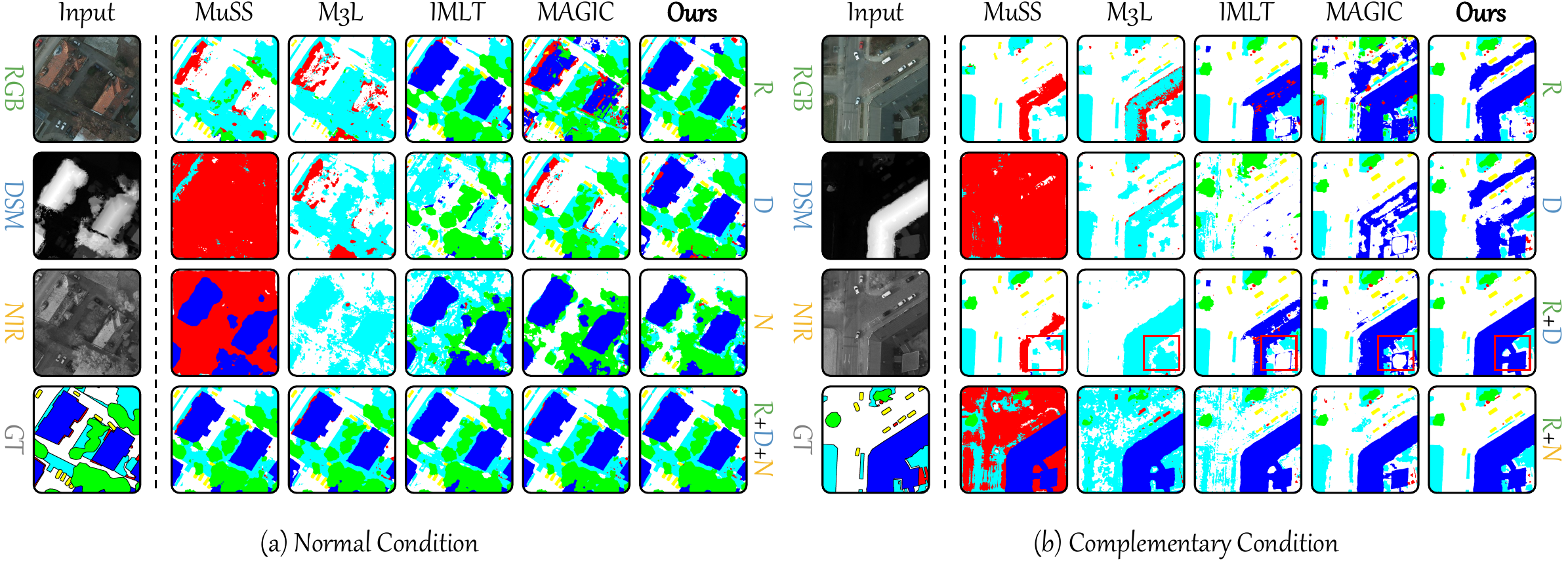}
		\caption{Qualitative comparisons on the \ISPRS{} under the (a) normal condition and (b) complementary condition.}
		\label{fig: quantitative_experiments_isprs}
		
		
		
	\end{figure*}
	
	\begin{figure*}[htb]
		\centering
		\includegraphics[width=1\linewidth]{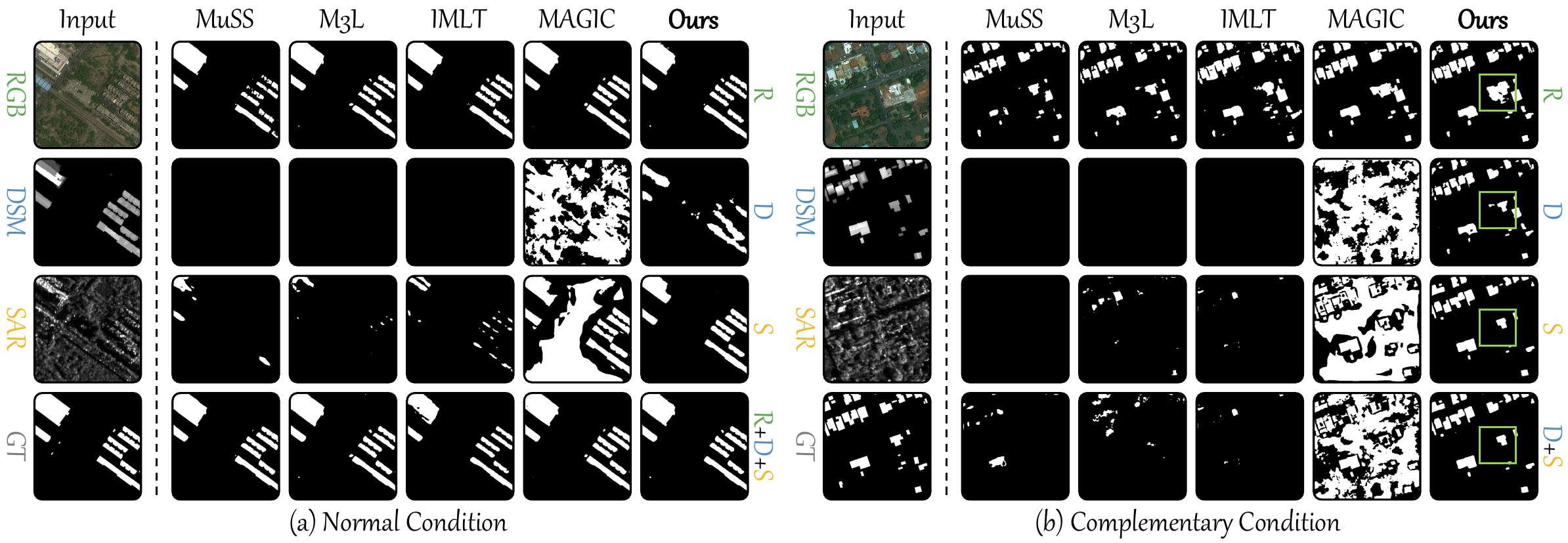}
		\caption{Qualitative comparisons on the \DFC{} under the (a) normal condition and (b) complementary condition.}
		
		\label{fig: quantitative_experiments_dfc}
		
		
	\end{figure*}
	
	\begin{figure*}[htb]
		\centering
		\includegraphics[width=1\linewidth]{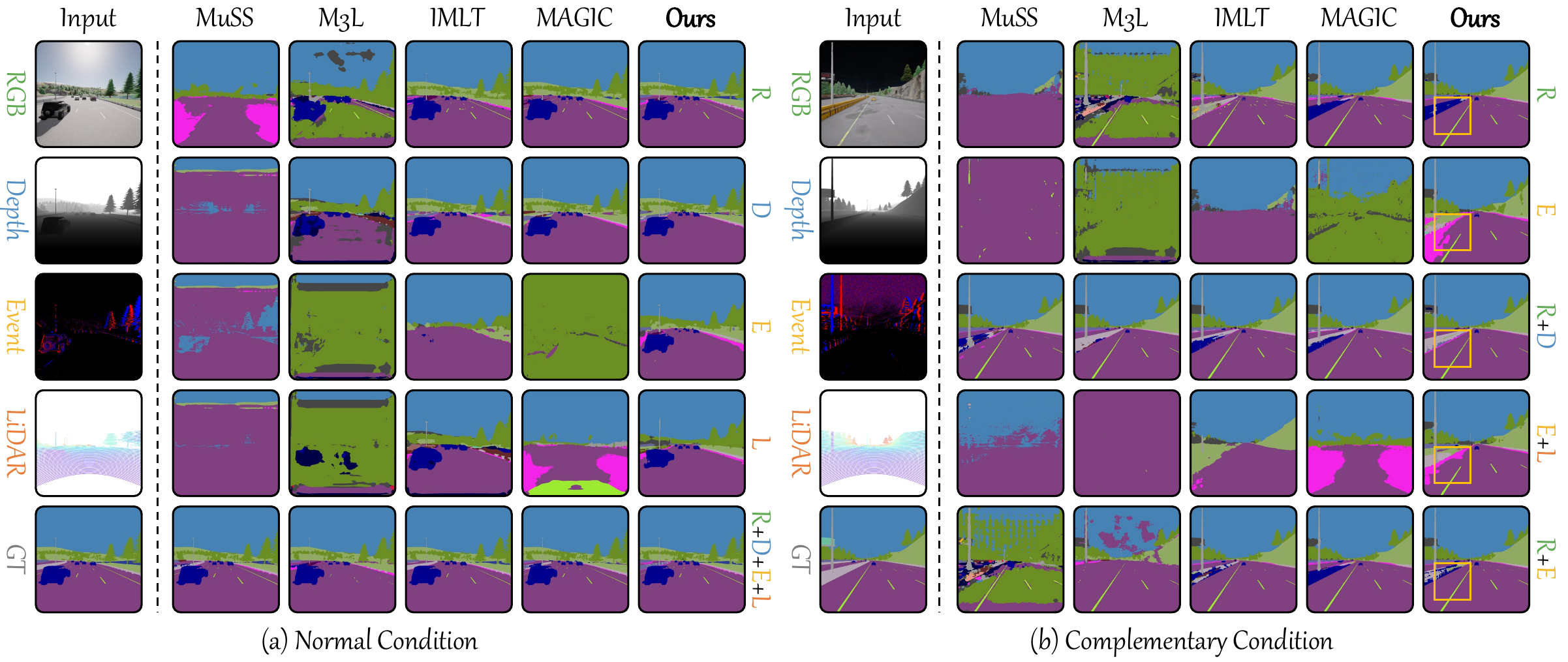}
		\caption{Qualitative comparisons on the \DELIVER{}  under the (a) normal condition and (b) complementary condition.}
		
		\label{fig: quantitative_experiments_deliver}
		
		
	\end{figure*}

	\subsubsection{Metircs}
	
	We evaluate segmentation performance using both mean Intersection over Union (mIoU) and F1-score (F1). \textit{Average}, \textit{Top-1}, and \textit{Last-1} denote the average, best, and worst performance across all-modal combinations, which assess overall effectiveness, optimal complementary effects, and robustness to fragile modalities, respectively. These metrics provide a comprehensive evaluation of the proposed SGMA's any-modal segmentation performance compared to state-of-the-art methods.
	
	\subsubsection{Implementation Details}
	
	Experiments on all datasets are conducted on 4 NVIDIA A100 GPUs. The initial learning rate is set to $6\times10^{-5}$, with polynomial decay (power of 0.9) over 200 epochs. A 10-epoch warm-up at 10\% of the initial learning rate is applied to stabilize training. We use AdamW optimizer with epsilon $10^{-8}$, weight decay $10^{-2}$. In the loss function, the weighting coefficients $\lambda_{\text{SGF}}$ and $\lambda_{\text{MAS}}$ are set to 2 and 1, respectively.
	
	\subsection{Qualitative Comparisons}
	
	We conduct qualitative evaluations under two representative scenarios: (1) \textit{Normal Conditions} refer to single-modality configurations and full-modality combinations; (2) \textit{Complementary Conditions} involve modality-pair configurations designed to exploit complementary properties. As illustrated in \cref{fig: quantitative_experiments_isprs}, \cref{fig: quantitative_experiments_dfc}, and \cref{fig: quantitative_experiments_deliver}, these scenarios assess the robustness and adaptability of our method across diverse modality settings.
	
	\subsubsection{Normal Conditions}
	For robust modalities such as RGB, nearly all methods achieve satisfactory segmentation results, as shown in the first rows of \cref{fig: quantitative_experiments_isprs} and \cref{fig: quantitative_experiments_dfc}(a). However, fragile modalities cause baseline methods to fail severely. DSM on \ISPRS{} and \DFC{}, SAR on \DFC{}, and Event and LiDAR on \DELIVER{} all produce fragmented or nearly empty predictions in baseline methods. Our method maintains meaningful segmentation with recognizable structures across all these challenging conditions. Under full-modality settings, our method delivers the most complete segmentation across all datasets, demonstrating superior delineation of \textit{vegetation}, \textit{buildings}, pedestrians, vehicles, and road markings, as shown in the bottom rows of \cref{fig: quantitative_experiments_isprs}, \cref{fig: quantitative_experiments_dfc}, and \cref{fig: quantitative_experiments_deliver}(a).
	
	\subsubsection{Complementary Conditions}
	In robust+robust configurations, our method achieves more complete segmentation with better fine-grained detail preservation. In challenging robust+fragile combinations, our method leverages complementary information most effectively. RGB+DSM on both \ISPRS{} and \DFC{} successfully integrates elevation cues to produce precise \textit{building} boundaries, while RGB+Event on \DELIVER{} exploits motion sensitivity for enhanced dynamic object segmentation. Most remarkably, our method uniquely succeeds in fragile+fragile combinations where baseline methods fail catastrophically. DSM+SAR on \DFC{} and Event+LiDAR on \DELIVER{} produce coherent segmentation that even surpasses single robust modality performance, as highlighted in \cref{fig: quantitative_experiments_dfc} and \cref{fig: quantitative_experiments_deliver}(b). This demonstrates our superior capability in integrating complementary cues, validating the generalizability of our framework across remote sensing and autonomous driving domains.
	
	Across all three datasets, baseline methods frequently yield incomplete or erroneous segmentation under challenging modality combinations. Our method consistently maintains robust performance, particularly excelling in scenarios involving fragile modalities and their combinations. The consistent improvements across domains confirm our framework's broad applicability to diverse multimodal segmentation tasks.

	\subsection{Quantitative Comparisons}
	
	\begin{table*}[htb]
		\centering
		\caption{Mean Intersection over Union (\%) on the \ISPRS{} across various modality combinations and backbones. \textit{Average}, \textit{Top-1}, and \textit{Last-1} indicate the average, highest, and lowest mIoU among all arbitrary-modal combinations, respectively. \textbf{Bold} and \uline{underlined} values mark the best and second-best performances in each column. The modalities are denoted as follows: RGB (R), DSM (D), and NIR (N).}
			\begin{tabular}{c|l|cccccccccc}
\hline
\multicolumn{1}{l|}{Backbones} & Methods & R & D & N & R+D & R+N & D+N & R+D+N & \textit{Average} & \textit{Top-1} & \textit{Last-1} \\ \hline
 & \MuSS{} & 40.21 & 17.13 & 1.36 & {\ul 83.75} & 57.71 & 31.52 & {\ul 86.50} & 45.45 & {\ul 86.50} & 1.36 \\
 & \MtL{} & 30.72 & 10.41 & 20.99 & 81.31 & 78.54 & 72.76 & 84.07 & 54.12 & 84.07 & 10.41 \\
 & \IMLT{} & 69.57 & {\ul 38.78} & {\ul 69.82} & 67.82 & {\ul 80.03} & {\ul 81.29} & 85.12 & {\ul 70.35} & 85.12 & {\ul 38.78} \\
 & \MAGIC{} & {\ul 81.39} & 34.34 & 46.97 & 83.27 & 77.99 & 63.30 & 84.75 & 67.43 & 84.75 & 34.34 \\
 & SGMA   (\textbf{Ours}) & \textbf{83.51} & \textbf{57.05} & \textbf{76.06} & \textbf{86.62} & \textbf{84.25} & \textbf{82.56} & \textbf{86.84} & \textbf{79.55} & \textbf{86.84} & \textbf{57.05} \\
\multirow{-6}{*}{Pvt-v2-b2} & \cellcolor[HTML]{D9D9D9}\wrtSOTA{} & \cellcolor[HTML]{D9D9D9}+2.12 & \cellcolor[HTML]{D9D9D9}+18.26 & \cellcolor[HTML]{D9D9D9}+6.24 & \cellcolor[HTML]{D9D9D9}+2.87 & \cellcolor[HTML]{D9D9D9}+4.22 & \cellcolor[HTML]{D9D9D9}+1.26 & \cellcolor[HTML]{D9D9D9}+0.34 & \cellcolor[HTML]{D9D9D9}+9.20 & \cellcolor[HTML]{D9D9D9}+0.34 & \cellcolor[HTML]{D9D9D9}+18.26 \\ \hline
 & \MuSS{} & 20.75 & 6.75 & 6.83 & 43.25 & 51.49 & 20.61 & {\ul 84.83} & 33.50 & {\ul 84.83} & 6.75 \\
 & \MtL{} & 22.80 & 9.33 & 18.51 & 77.81 & 69.95 & {\ul 68.17} & 84.04 & 50.09 & 84.04 & 9.33 \\
 & \IMLT{} & {\ul 80.47} & {\ul 36.67} & 26.51 & {\ul 84.06} & {\ul 82.08} & 45.16 & 84.33 & 62.75 & 84.33 & 26.51 \\
 & \MAGIC{} & 75.85 & 32.02 & {\ul 54.98} & 78.93 & 77.84 & 60.21 & 83.65 & {\ul 66.21} & 83.65 & {\ul 32.02} \\
 & SGMA   (\textbf{Ours}) & \textbf{82.74} & \textbf{50.57} & \textbf{71.17} & \textbf{84.63} & \textbf{83.28} & \textbf{77.00} & \textbf{85.55} & \textbf{76.42} & \textbf{85.55} & \textbf{50.57} \\
\multirow{-6}{*}{ResNet-50} & \cellcolor[HTML]{D9D9D9}\wrtSOTA{} & \cellcolor[HTML]{D9D9D9}+2.27 & \cellcolor[HTML]{D9D9D9}+13.90 & \cellcolor[HTML]{D9D9D9}+16.19 & \cellcolor[HTML]{D9D9D9}+0.57 & \cellcolor[HTML]{D9D9D9}+1.20 & \cellcolor[HTML]{D9D9D9}+8.83 & \cellcolor[HTML]{D9D9D9}+0.72 & \cellcolor[HTML]{D9D9D9}+10.21 & \cellcolor[HTML]{D9D9D9}+0.72 & \cellcolor[HTML]{D9D9D9}+18.54 \\ \hline
\end{tabular}
		\label{tab: isprs_miou}
	\end{table*}
	
	\begin{table*}[htb]
		\centering
		\caption{F1-score (\%) on the \ISPRS{} across various modality combinations and backbones. \textit{Average}, \textit{Top-1}, and \textit{Last-1} indicate the average, highest, and lowest F1-score among all arbitrary-modal combinations, respectively.}
			\begin{tabular}{c|l|cccccccccc}
\hline
\multicolumn{1}{l|}{Backbones} & Methods & R & D & N & R+D & R+N & D+N & R+D+N & \textit{Average} & \textit{Top-1} & \textit{Last-1} \\ \hline
 & \MuSS{} & 51.96 & 19.44 & 2.63 & {\ul 90.95} & 67.42 & 40.88 & {\ul 92.61} & 52.27 & {\ul 92.61} & 2.63 \\
 & \MtL{} & 41.27 & 14.79 & 29.34 & 89.51 & 87.90 & 84.02 & 90.36 & 62.45 & 90.36 & 14.79 \\
 & \IMLT{} & 80.55 & {\ul 48.59} & {\ul 81.95} & 80.07 & {\ul 88.84} & {\ul 89.40} & 91.38 & {\ul 80.11} & 91.38 & {\ul 48.59} \\
 & \MAGIC{} & {\ul 89.67} & 45.14 & 61.25 & 90.69 & 87.49 & 76.16 & 91.19 & 77.37 & 91.19 & 45.14 \\
 & SGMA (\textbf{Ours}) & \textbf{90.92} & \textbf{70.36} & \textbf{86.24} & \textbf{92.88} & \textbf{91.34} & \textbf{90.18} & \textbf{92.96} & \textbf{87.84} & \textbf{92.96} & \textbf{70.36} \\
\multirow{-6}{*}{Pvt-v2-b2} & \cellcolor[HTML]{D9D9D9}\wrtSOTA{} & \cellcolor[HTML]{D9D9D9}+1.25 & \cellcolor[HTML]{D9D9D9}+21.77 & \cellcolor[HTML]{D9D9D9}+4.29 & \cellcolor[HTML]{D9D9D9}+1.93 & \cellcolor[HTML]{D9D9D9}+2.50 & \cellcolor[HTML]{D9D9D9}+0.78 & \cellcolor[HTML]{D9D9D9}+0.35 & \cellcolor[HTML]{D9D9D9}+7.73 & \cellcolor[HTML]{D9D9D9}+0.35 & \cellcolor[HTML]{D9D9D9}+21.77 \\ \hline
 & \MuSS{} & 27.28 & 10.10 & 10.33 & 54.51 & 65.14 & 28.07 & {\ul 91.65} & 41.01 & {\ul 91.65} & 10.10 \\
 & \MtL{} & 31.61 & 13.43 & 26.36 & 87.36 & 82.23 & {\ul 80.80} & 90.76 & 58.94 & 90.76 & 13.43 \\
 & \IMLT{} & {\ul 89.07} & {\ul 50.02} & 41.37 & {\ul 90.74} & {\ul 90.04} & 59.45 & 91.20 & 73.13 & 91.20 & 41.37 \\
 & \MAGIC{} & 86.22 & 42.58 & {\ul 68.69} & 87.87 & 87.48 & 73.26 & 90.94 & {\ul 76.72} & 90.94 & {\ul 42.58} \\
 & SGMA (\textbf{Ours}) & \textbf{90.41} & \textbf{64.17} & \textbf{82.88} & \textbf{91.52} & \textbf{90.74} & \textbf{86.48} & \textbf{92.06} & \textbf{85.47} & \textbf{92.06} & \textbf{64.17} \\
\multirow{-6}{*}{ResNet-50} & \cellcolor[HTML]{D9D9D9}\wrtSOTA{} & \cellcolor[HTML]{D9D9D9}+1.34 & \cellcolor[HTML]{D9D9D9}+14.16 & \cellcolor[HTML]{D9D9D9}+14.19 & \cellcolor[HTML]{D9D9D9}+0.78 & \cellcolor[HTML]{D9D9D9}+0.70 & \cellcolor[HTML]{D9D9D9}+5.68 & \cellcolor[HTML]{D9D9D9}+0.41 & \cellcolor[HTML]{D9D9D9}+8.75 & \cellcolor[HTML]{D9D9D9}+0.41 & \cellcolor[HTML]{D9D9D9}+21.59 \\ \hline
\end{tabular}
		\label{tab: isprs_f1}
	\end{table*}
	
	\begin{table*}[htb]
		\centering
		\caption{Mean Intersection over Union  (\%) on the \DFC{} across various modality combinations and backbones. \textit{Average}, \textit{Top-1}, and \textit{Last-1} indicate the average, highest, and lowest mIoU among all arbitrary-modal combinations, respectively. The modalities are denoted as follows: RGB (R), DSM (D), and SAR (S).}
			\begin{tabular}{c|l|cccccccccc}
\hline
\multicolumn{1}{l|}{Backbones} & Methods & R & D & S & R+D & R+S & D+S & R+D+S & \textit{Average} & \textit{Top-1} & \textit{Last-1} \\ \hline
 & \MuSS{} & 88.12 & 52.68 & 37.51 & {\ul 91.28} & 87.09 & 49.70 & {\ul 92.21} & 71.23 & {\ul 92.21} & 37.51 \\
 & \MtL{} & {\ul 90.55} & 50.45 & {\ul 37.59} & 88.88 & 90.20 & {\ul 71.01} & 91.10 & {\ul 74.25} & 91.10 & {\ul 37.59} \\
 & \IMLT{} & 90.54 & 53.73 & 32.53 & 90.81 & {\ul 90.61} & 49.98 & 91.12 & 71.33 & 91.12 & 32.53 \\
 & \MAGIC{} & 88.98 & {\ul 65.96} & 37.51 & 89.20 & 83.29 & 43.75 & 81.93 & 70.09 & 89.20 & 37.51 \\
 & SGMA (\textbf{Ours}) & \textbf{90.84} & \textbf{76.70} & \textbf{53.13} & \textbf{91.95} & \textbf{90.98} & \textbf{77.47} & \textbf{92.29} & \textbf{81.91} & \textbf{92.29} & \textbf{53.13} \\
\multirow{-6}{*}{Pvt-v2-b2} & \cellcolor[HTML]{D9D9D9}\wrtSOTA{} & \cellcolor[HTML]{D9D9D9}+0.29 & \cellcolor[HTML]{D9D9D9}+10.74 & \cellcolor[HTML]{D9D9D9}+15.54 & \cellcolor[HTML]{D9D9D9}+0.67 & \cellcolor[HTML]{D9D9D9}+0.37 & \cellcolor[HTML]{D9D9D9}+6.47 & \cellcolor[HTML]{D9D9D9}+0.08 & \cellcolor[HTML]{D9D9D9}+7.66 & \cellcolor[HTML]{D9D9D9}+0.08 & \cellcolor[HTML]{D9D9D9}+15.54 \\ \hline
 & \MuSS{} & 61.02 & 51.49 & 37.52 & 86.57 & 78.11 & 69.68 & {\ul 89.63} & 67.72 & {\ul 89.63} & 37.52 \\
 & \MtL{} & 73.12 & 54.09 & 37.57 & 87.65 & 84.88 & {\ul 71.45} & 88.73 & 71.07 & 88.73 & 37.57 \\
 & \IMLT{} & {\ul 86.84} & 65.49 & {\ul 43.37} & {\ul 88.46} & {\ul 87.29} & 63.48 & 88.62 & 74.79 & 88.62 & {\ul 43.37} \\
 & \MAGIC{} & 84.23 & {\ul 73.53} & 40.78 & 87.43 & 83.15 & 69.93 & 86.55 & {\ul 75.09} & 87.43 & 40.78 \\
 & SGMA (\textbf{Ours}) & \textbf{87.80} & \textbf{76.91} & \textbf{49.71} & \textbf{89.52} & \textbf{88.05} & \textbf{78.00} & \textbf{90.57} & \textbf{80.08} & \textbf{90.57} & \textbf{49.71} \\
\multirow{-6}{*}{ResNet-50} & \cellcolor[HTML]{D9D9D9}\wrtSOTA{} & \cellcolor[HTML]{D9D9D9}+0.96 & \cellcolor[HTML]{D9D9D9}+3.38 & \cellcolor[HTML]{D9D9D9}+6.34 & \cellcolor[HTML]{D9D9D9}+1.06 & \cellcolor[HTML]{D9D9D9}+0.77 & \cellcolor[HTML]{D9D9D9}+6.55 & \cellcolor[HTML]{D9D9D9}+0.94 & \cellcolor[HTML]{D9D9D9}+4.99 & \cellcolor[HTML]{D9D9D9}+0.94 & \cellcolor[HTML]{D9D9D9}+6.34 \\ \hline
\end{tabular}
		\label{tab: dfc_miou}
	\end{table*}
	
	\begin{table*}[htb]
		\centering
		\caption{F1-score (\%) on the \DFC{} across various modality combinations and backbones. \textit{Average}, \textit{Top-1}, and \textit{Last-1} indicate the average, highest, and lowest F1-score among all arbitrary-modal combinations, respectively.}
			\begin{tabular}{c|l|cccccccccc}
\hline
\multicolumn{1}{l|}{Backbones} & Methods & R & D & S & R+D & R+S & D+S & R+D+S & \textit{Average} & \textit{Top-1} & \textit{Last-1} \\ \hline
 & \MuSS{} & 93.58 & 64.78 & 42.86 & {\ul 95.39} & 92.97 & 61.18 & {\ul 95.40} & 78.02 & {\ul 95.40} & 42.86 \\
 & \MtL{} & 94.14 & 62.07 & 43.00 & 94.02 & 93.78 & {\ul 80.55} & 94.29 & {\ul 80.27} & 94.29 & 43.00 \\
 & \IMLT{} & {\ul 94.53} & 69.01 & {\ul 48.64} & 94.69 & {\ul 94.61} & 65.77 & 94.75 & 80.29 & 94.75 & {\ul 48.64} \\
 & \MAGIC{} & 94.09 & {\ul 78.06} & 42.86 & 94.21 & 90.64 & 53.05 & 89.76 & 77.52 & 94.21 & 42.86 \\
 & SGMA (\textbf{Ours}) & \textbf{94.98} & \textbf{86.39} & \textbf{66.28} & \textbf{95.76} & \textbf{95.02} & \textbf{86.93} & \textbf{95.91} & \textbf{88.75} & \textbf{95.91} & \textbf{66.28} \\
\multirow{-6}{*}{Pvt-v2-b2} & \cellcolor[HTML]{D9D9D9}\wrtSOTA{} & \cellcolor[HTML]{D9D9D9}+0.45 & \cellcolor[HTML]{D9D9D9}+8.33 & \cellcolor[HTML]{D9D9D9}+17.64 & \cellcolor[HTML]{D9D9D9}+0.37 & \cellcolor[HTML]{D9D9D9}+0.40 & \cellcolor[HTML]{D9D9D9}+6.38 & \cellcolor[HTML]{D9D9D9}+0.51 & \cellcolor[HTML]{D9D9D9}+8.46 & \cellcolor[HTML]{D9D9D9}+0.51 & \cellcolor[HTML]{D9D9D9}+17.64 \\ \hline
 & \MuSS{} & 73.51 & 63.46 & 42.88 & 92.67 & 87.29 & 81.20 & {\ul 94.46} & 76.50 & {\ul 94.46} & 42.88 \\
 & \MtL{} & 83.67 & 66.38 & 42.97 & {\ul 93.32} & 91.65 & {\ul 80.21} & 93.94 & 78.88 & 93.94 & 42.97 \\
 & \IMLT{} & {\ul 92.56} & 79.82 & {\ul 59.66} & 92.93 & {\ul 92.70} & 78.16 & 93.05 & {\ul 84.13} & 93.05 & {\ul 59.66} \\
 & \MAGIC{} & 91.27 & {\ul 84.18} & 50.33 & 91.35 & 90.58 & 81.36 & 91.83 & 82.99 & 91.83 & 50.33 \\
 & SGMA (\textbf{Ours}) & \textbf{92.83} & \textbf{86.54} & \textbf{62.14} & \textbf{94.40} & \textbf{93.10} & \textbf{87.28} & \textbf{94.99} & \textbf{87.32} & \textbf{94.99} & \textbf{62.14} \\
\multirow{-6}{*}{ResNet-50} & \cellcolor[HTML]{D9D9D9}\wrtSOTA{} & \cellcolor[HTML]{D9D9D9}+0.28 & \cellcolor[HTML]{D9D9D9}+2.36 & \cellcolor[HTML]{D9D9D9}+2.48 & \cellcolor[HTML]{D9D9D9}+1.08 & \cellcolor[HTML]{D9D9D9}+0.40 & \cellcolor[HTML]{D9D9D9}+5.92 & \cellcolor[HTML]{D9D9D9}+0.53 & \cellcolor[HTML]{D9D9D9}+3.20 & \cellcolor[HTML]{D9D9D9}+0.53 & \cellcolor[HTML]{D9D9D9}+2.48 \\ \hline
\end{tabular}
		\label{tab: dfc_f1}
	\end{table*}

	\begin{table*}[htb]
		\centering
		\caption{Mean Intersection over Union (\%) on the \DELIVER{} across various modality combinations and backbones. \textit{Average}, \textit{Top-1}, and \textit{Last-1} indicate the average, highest, and lowest mIoU among all arbitrary-modal combinations, respectively. The modalities are denoted as follows: RGB (R), Depth (D), Event (E), and LiDAR (L).}
		\label{tab: deliver_miou}
			\begin{tabular}{c|l|ccccccccc}
\hline
\multicolumn{1}{l|}{Backbones} & Methods & R & D & E & L & RD & RE & RL & DE & DL \\ \hline
 & \MuSS & 3.90 & 0.70 & 1.15 & 0.54 & 50.15 & 13.12 & 18.22 & 21.46 & 4.01 \\
 & \MtL & 18.67 & 37.45 & 2.12 & 1.33 & 59.78 & 24.23 & 27.34 & 41.56 & 34.12 \\
 & \IMLT & {\ul 38.45} & 52.89 & {\ul 17.52} & {\ul 19.67} & 61.89 & {\ul 40.78} & {\ul 41.45} & {\ul 58.67} & {\ul 58.01} \\
 & \MAGIC & 29.96 & {\ul 55.80} & 3.74 & 4.62 & {\ul 64.01} & 30.12 & 32.83 & 56.71 & 56.07 \\
 & SGMA   (\textbf{Ours}) & \textbf{55.12} & \textbf{58.04} & \textbf{30.87} & \textbf{29.22} & \textbf{67.60} & \textbf{55.69} & \textbf{56.16} & \textbf{61.40} & \textbf{59.95} \\
 & \cellcolor[HTML]{D9D9D9}\wrtSOTA & \cellcolor[HTML]{D9D9D9}+16.67 & \cellcolor[HTML]{D9D9D9}+2.24 & \cellcolor[HTML]{D9D9D9}+13.35 & \cellcolor[HTML]{D9D9D9}+9.55 & \cellcolor[HTML]{D9D9D9}+3.72 & \cellcolor[HTML]{D9D9D9}+14.91 & \cellcolor[HTML]{D9D9D9}+14.71 & \cellcolor[HTML]{D9D9D9}+2.73 & \cellcolor[HTML]{D9D9D9}+1.94 \\ \cline{2-11} 
 & Methods & EL & RDE & RDL & REL & DEL & RDEL & \textit{Average} & \textit{Top-1} & \textit{Last-1} \\ \cline{2-11}
 & \MuSS & 3.00 & \textbf{66.23} & {\ul 66.61} & 15.76 & 46.29 & {\ul 66.41} & 25.17 & {\ul 66.61} & 0.54 \\
 & \MtL & 4.45 & 62.89 & 63.13 & 29.56 & 53.67 & 62.45 & 34.85 & 63.13 & 1.33 \\
 & \IMLT & {\ul 22.89} & 63.23 & 63.56 & {\ul 44.12} & {\ul 60.45} & 64.04 & {\ul 47.17} & 64.04 & {\ul 17.52} \\
 & \MAGIC & 5.69 & 63.66 & 63.50 & 34.25 & 55.91 & 64.04 & 41.39 & 64.04 & 3.74 \\
 & SGMA   (\textbf{Ours}) & \textbf{38.46} & {\ul 66.01} & \textbf{67.69} & \textbf{56.52} & \textbf{61.82} & \textbf{67.73} & \textbf{55.49} & \textbf{67.73} & \textbf{29.22} \\
\multirow{-13}{*}{Pvt-v2-b2} & \cellcolor[HTML]{D9D9D9}\wrtSOTA & \cellcolor[HTML]{D9D9D9}+15.57 & \cellcolor[HTML]{D9D9D9}-0.22 & \cellcolor[HTML]{D9D9D9}+1.08 & \cellcolor[HTML]{D9D9D9}+12.40 & \cellcolor[HTML]{D9D9D9}+1.37 & \cellcolor[HTML]{D9D9D9}+1.19 & \cellcolor[HTML]{D9D9D9}+8.31 & \cellcolor[HTML]{D9D9D9}+1.12 & \cellcolor[HTML]{D9D9D9}+11.70 \\ \hline
\multicolumn{1}{l|}{Backbones} & Methods & R & D & E & L & RD & RE & RL & DE & DL \\ \hline
 & \MuSS & 0.91 & 0.34 & 0.57 & 0.39 & 46.93 & 9.82 & 13.73 & 2.58 & 1.71 \\
 & \MtL & 14.56 & 28.34 & 1.23 & 0.86 & 55.67 & 18.92 & 21.45 & 32.78 & 26.54 \\
 & \IMLT & \textbf{53.45} & {\ul 51.34} & {\ul 15.76} & {\ul 18.67} & {\ul 60.45} & {\ul 53.67} & {\ul 54.12} & {\ul 54.89} & {\ul 54.23} \\
 & \MAGIC & 45.75 & 41.91 & 1.50 & 6.58 & 58.43 & 45.61 & 45.77 & 45.07 & 45.15 \\
 & SGMA   (\textbf{Ours}) & {\ul 51.16} & \textbf{55.59} & \textbf{27.40} & \textbf{27.77} & \textbf{63.93} & \textbf{54.85} & \textbf{56.78} & \textbf{57.95} & \textbf{55.95} \\
 & \cellcolor[HTML]{D9D9D9}\wrtSOTA & \cellcolor[HTML]{D9D9D9}-2.29 & \cellcolor[HTML]{D9D9D9}+4.25 & \cellcolor[HTML]{D9D9D9}+11.64 & \cellcolor[HTML]{D9D9D9}+9.10 & \cellcolor[HTML]{D9D9D9}+3.48 & \cellcolor[HTML]{D9D9D9}+1.18 & \cellcolor[HTML]{D9D9D9}+2.66 & \cellcolor[HTML]{D9D9D9}+3.06 & \cellcolor[HTML]{D9D9D9}+1.72 \\ \cline{2-11} 
 & Methods & EL & RDE & RDL & REL & DEL & RDEL & \textit{Average} & \textit{Top-1} & \textit{Last-1} \\ \cline{2-11} 
 & \MuSS & 3.02 & {\ul 63.67} & {\ul 63.53} & 34.05 & \textbf{63.70} & {\ul 63.98} & 24.60 & {\ul 63.98} & 0.34 \\
 & \MtL & 3.45 & 60.45 & 60.78 & 23.67 & 48.23 & 62.11 & 30.60 & 62.11 & 0.86 \\
 & \IMLT & {\ul 20.45} & 61.89 & 62.23 & {\ul 52.05} & 56.67 & 62.65 & {\ul 48.83} & 62.65 & {\ul 15.76} \\
 & \MAGIC & 7.03 & 59.07 & 58.85 & 44.65 & 46.33 & 58.88 & 40.71 & 59.07 & 1.50 \\
 & SGMA   (\textbf{Ours}) & \textbf{35.61} & \textbf{64.17} & \textbf{64.31} & \textbf{55.34} & {\ul 63.66} & \textbf{64.79} & \textbf{53.28} & \textbf{64.79} & \textbf{27.40} \\
\multirow{-13}{*}{ResNet-50} & \cellcolor[HTML]{D9D9D9}\wrtSOTA & \cellcolor[HTML]{D9D9D9}+15.16 & \cellcolor[HTML]{D9D9D9}+0.50 & \cellcolor[HTML]{D9D9D9}+0.78 & \cellcolor[HTML]{D9D9D9}+3.29 & \cellcolor[HTML]{D9D9D9}-0.04 & \cellcolor[HTML]{D9D9D9}+0.81 & \cellcolor[HTML]{D9D9D9}+4.45 & \cellcolor[HTML]{D9D9D9}+0.81 & \cellcolor[HTML]{D9D9D9}+11.64 \\ \hline
\end{tabular}
	\end{table*}
	
	\begin{table*}[htb]
		\centering
		\caption{F1-scor (\%) on the \DELIVER{} across various modality combinations and backbones. \textit{Average}, \textit{Top-1}, and \textit{Last-1} indicate the average, highest, and lowest mIoU among all arbitrary-modal combinations, respectively.}
		\label{tab: deliver_f1}
			\begin{tabular}{c|l|ccccccccc}
\hline
\multicolumn{1}{l|}{Backbones} & Methods & R & D & E & L & RD & RE & RL & DE & DL \\ \hline
 & \MuSS & 7.51 & 1.39 & 2.27 & 1.07 & 66.80 & 23.20 & 30.82 & 35.34 & 7.71 \\
 & \MtL & 31.47 & 54.49 & 4.15 & 2.63 & 74.83 & 39.01 & 42.94 & 58.72 & 50.88 \\
 & \IMLT & {\ul 55.54} & 69.19 & {\ul 29.82} & {\ul 32.87} & 76.46 & {\ul 57.93} & {\ul 58.61} & {\ul 73.95} & {\ul 73.43} \\
 & \MAGIC & 46.11 & {\ul 71.63} & 7.21 & 8.83 & {\ul 78.06} & 46.29 & 49.43 & 72.38 & 71.85 \\
 & SGMA   (\textbf{Ours}) & \textbf{71.07} & \textbf{73.45} & \textbf{47.18} & \textbf{45.22} & \textbf{80.67} & \textbf{71.54} & \textbf{71.93} & \textbf{76.08} & \textbf{74.96} \\
 & \cellcolor[HTML]{D9D9D9}\wrtSOTA & \cellcolor[HTML]{D9D9D9}+15.53 & \cellcolor[HTML]{D9D9D9}+1.82 & \cellcolor[HTML]{D9D9D9}+17.36 & \cellcolor[HTML]{D9D9D9}+12.35 & \cellcolor[HTML]{D9D9D9}+2.70 & \cellcolor[HTML]{D9D9D9}+13.61 & \cellcolor[HTML]{D9D9D9}+13.32 & \cellcolor[HTML]{D9D9D9}+2.13 & \cellcolor[HTML]{D9D9D9}+1.53 \\ \cline{2-11} 
 & Methods & EL & RDE & RDL & REL & DEL & RDEL & \textit{Average} & \textit{Top-1} & \textit{Last-1} \\ \cline{2-11}
 & \MuSS & 5.83 & \textbf{79.69} & {\ul 79.96} & 27.23 & 63.28 & {\ul 79.82} & 34.13 & {\ul 79.96} & 1.07 \\
 & \MtL & 8.52 & 77.22 & 77.40 & 45.63 & 69.85 & 76.89 & 47.64 & 77.40 & 2.63 \\
 & \IMLT & {\ul 37.25} & 77.47 & 77.72 & {\ul 61.23} & {\ul 75.35} & 78.08 & {\ul 62.33} & 78.08 & {\ul 29.82} \\
 & \MAGIC & 10.77 & 77.80 & 77.68 & 51.02 & 71.72 & 78.08 & 54.59 & 78.08 & 7.21 \\
 & SGMA   (\textbf{Ours}) & \textbf{55.55} & {\ul 79.53} & \textbf{80.73} & \textbf{72.22} & \textbf{76.41} & \textbf{80.76} & \textbf{70.49} & \textbf{80.76} & \textbf{45.22} \\
\multirow{-13}{*}{Pvt-v2-b2} & \cellcolor[HTML]{D9D9D9}\wrtSOTA & \cellcolor[HTML]{D9D9D9}+18.30 & \cellcolor[HTML]{D9D9D9}-0.16 & \cellcolor[HTML]{D9D9D9}+0.77 & \cellcolor[HTML]{D9D9D9}+10.99 & \cellcolor[HTML]{D9D9D9}+1.06 & \cellcolor[HTML]{D9D9D9}+0.85 & \cellcolor[HTML]{D9D9D9}+8.16 & \cellcolor[HTML]{D9D9D9}+0.80 & \cellcolor[HTML]{D9D9D9}+15.40 \\ \hline
\multicolumn{1}{l|}{Backbones} & Methods & R & D & E & L & RD & RE & RL & DE & DL \\ \hline
 & \MuSS & 1.80 & 0.68 & 1.13 & 0.78 & 63.88 & 17.88 & 24.15 & 5.03 & 3.36 \\
 & \MtL & 25.42 & 44.16 & 2.43 & 1.71 & 71.52 & 31.82 & 35.32 & 49.37 & 41.95 \\
 & \IMLT & \textbf{69.66} & {\ul 67.85} & {\ul 27.23} & {\ul 31.46} & {\ul 75.35} & {\ul 69.85} & {\ul 70.23} & {\ul 70.88} & {\ul 70.32} \\
 & \MAGIC & 62.78 & 59.06 & 2.96 & 12.35 & 73.76 & 62.65 & 62.80 & 62.14 & 62.21 \\
 & SGMA   (\textbf{Ours}) & {\ul 67.69} & \textbf{71.46} & \textbf{43.01} & \textbf{43.47} & \textbf{78.00} & \textbf{70.84} & \textbf{72.43} & \textbf{73.38} & \textbf{71.75} \\
 & \cellcolor[HTML]{D9D9D9}\wrtSOTA & \cellcolor[HTML]{D9D9D9}-1.97 & \cellcolor[HTML]{D9D9D9}+3.61 & \cellcolor[HTML]{D9D9D9}+15.78 & \cellcolor[HTML]{D9D9D9}+12.01 & \cellcolor[HTML]{D9D9D9}+2.65 & \cellcolor[HTML]{D9D9D9}+0.99 & \cellcolor[HTML]{D9D9D9}+2.20 & \cellcolor[HTML]{D9D9D9}+2.50 & \cellcolor[HTML]{D9D9D9}+1.43 \\ \cline{2-11} 
 & Methods & EL & RDE & RDL & REL & DEL & RDEL & \textit{Average} & \textit{Top-1} & \textit{Last-1} \\ \cline{2-11} 
 & \MuSS & 5.86 & {\ul 77.80} & {\ul 77.70} & 50.80 & \textbf{77.83} & {\ul 78.03} & 32.45 & {\ul 78.03} & 0.68 \\
 & \MtL & 6.67 & 75.35 & 75.61 & 38.28 & 65.07 & 76.63 & 42.75 & 76.63 & 1.71 \\
 & \IMLT & {\ul 33.96} & 76.46 & 76.72 & {\ul 68.46} & 72.34 & 77.04 & {\ul 63.85} & 77.04 & {\ul 27.23} \\
 & \MAGIC & 13.14 & 74.27 & 74.10 & 61.74 & 63.32 & 74.12 & 54.76 & 74.27 & 2.96 \\
 & SGMA   (\textbf{Ours}) & \textbf{52.52} & \textbf{78.18} & \textbf{78.28} & \textbf{71.25} & {\ul 77.80} & \textbf{78.63} & \textbf{68.58} & \textbf{78.63} & \textbf{43.01} \\
\multirow{-13}{*}{ResNet-50} & \cellcolor[HTML]{D9D9D9}\wrtSOTA & \cellcolor[HTML]{D9D9D9}+18.56 & \cellcolor[HTML]{D9D9D9}+0.38 & \cellcolor[HTML]{D9D9D9}+0.58 & \cellcolor[HTML]{D9D9D9}+2.79 & \cellcolor[HTML]{D9D9D9}-0.03 & \cellcolor[HTML]{D9D9D9}+0.60 & \cellcolor[HTML]{D9D9D9}+4.73 & \cellcolor[HTML]{D9D9D9}+0.60 & \cellcolor[HTML]{D9D9D9}+15.78 \\ \hline
\end{tabular}
	\end{table*}

	We conduct quantitative comparisons across \ISPRS{}, \DFC{}, and \DELIVER{} to evaluate the effectiveness of our method with arbitrary modality combinations and different backbones. The quantitative results present mIoU and F1-score metrics, where \cref{tab: isprs_miou}, \cref{tab: isprs_f1}, \cref{tab: dfc_miou}, \cref{tab: dfc_f1}, \cref{tab: deliver_miou}, and \cref{tab: deliver_f1} report results on all three datasets.
	
	\subsubsection{Scalability with Increasing Modalities}
	
	Our method demonstrates substantial improvements over SOTA baselines across all three datasets. With PVT-v2-b2, we achieve +9.20\%, +7.66\%, and +8.31\% in \textit{Average} mIoU on \ISPRS{}, \DFC{}, and \DELIVER{}, respectively, as shown in \cref{tab: isprs_miou}, \cref{tab: dfc_miou}, and \cref{tab: deliver_miou}. The corresponding F1-score improvements are +7.73\%, +8.46\%, and +8.16\%, demonstrating robust generalization from remote sensing to complex urban scenarios. More importantly, our method excels in the most challenging single-modality scenarios, achieving \textit{Last-1} performance gains of +18.26\%, +15.54\%, and +11.70\% in mIoU across the three datasets. These consistent improvements validate the effectiveness of our framework in handling fragile modalities, regardless of sensor type.
	
	\subsubsection{Average Performance Across Modalities}
	
	A distinctive advantage of our method is its consistent performance improvement as modality count increases, demonstrating effective multi-modal complementarity. In contrast, baseline methods often suffer from performance degradation when additional modalities are introduced. As evidenced in \cref{tab: isprs_miou} and \cref{tab: deliver_miou}, MAGIC drops by 3.40\% when moving from RGB to RGB+NIR on \ISPRS{}, while IMLT experiences 0.52\% degradation when transitioning to full-modality on \DELIVER{}. Our method consistently maintains or improves performance with additional modalities across all datasets, achieving +1.12\% gain from RGB+Depth+LiDAR to full-modality on \DELIVER{}. This validates that our framework effectively harmonizes cross-modal heterogeneity, enabling robust fusion regardless of modality combinations or sensor configurations.
	
	\subsubsection{Robustness with Fragile Modalities}
	
	Our method demonstrates exceptional capability in handling fragile modalities and their combinations across different domains. On single fragile modalities, we achieve +18.26\%, +15.54\%, and +13.35\% improvements over SOTA on DSM in \ISPRS{}, SAR in \DFC{}, and Event in \DELIVER{}, respectively, as shown in \cref{tab: isprs_miou}, \cref{tab: dfc_miou}, and \cref{tab: deliver_miou}. In challenging fragile+fragile combinations, our method delivers remarkable gains: DSM+SAR achieves +6.47\% improvement in \textit{Average} on \DFC{}, while Event+LiDAR achieves +15.57\% on \DELIVER{}. This validates our framework's unique ability to leverage modality-specific strengths across heterogeneous sensor types, confirming broad applicability across remote sensing and autonomous driving IMSS tasks.
	
	\subsubsection{Cross-Backbone Generalization}
	
	Our SGMA framework exhibits plug-and-play characteristics, enabling seamless integration across different backbone architectures. With ResNet-50, SGMA delivers consistent improvements across all datasets: +10.21\%, +4.99\%, and +4.45\% in \textit{Average} mIoU on \ISPRS{}, \DFC{}, and \DELIVER{}, respectively, while maintaining strong \textit{Last-1} gains of +18.54\%, +6.34\%, and +11.64\%, as shown in \cref{tab: isprs_miou}, \cref{tab: dfc_miou}, and \cref{tab: deliver_miou}. The performance improvements with ResNet-50 follow consistent trends observed with PVT-v2-b2 across all three datasets. This consistency confirms our framework's architecture-agnostic nature and practical applicability across diverse multimodal segmentation tasks in both remote sensing and autonomous driving domains.
	
	\subsection{Ablation Studies}
	
	\begin{table*}[htb]
		\centering
		\caption{Mean Intersection over Union (\%) on the \ISPRS{} for Ablation Studies of SGF and MAS Modules. Each \textbf{Variant} denotes a different component configuration. \textit{Average}, \textit{Top-1}, and \textit{Last-1} indicate the average, highest, and lowest mIoU among all arbitrary-modal combinations, respectively. The modalities are denoted as follows: RGB (R), DSM (D), and SAR (S). The red arrows and values indicate the performance drop compared to \textbf{Variant (c)}, while the green arrows denote the improvement.}
		\resizebox{\linewidth}{!}{
			\begin{tabular}{c|cc|cccccccccc}
\hline
\multirow{2}{*}{Variants} & \multicolumn{2}{c|}{Components} & \multirow{2}{*}{R} & \multirow{2}{*}{D} & \multirow{2}{*}{N} & \multirow{2}{*}{RD} & \multirow{2}{*}{RN} & \multirow{2}{*}{DN} & \multirow{2}{*}{RDN} & \multirow{2}{*}{\textit{Average}} & \multirow{2}{*}{\textit{Top-1}} & \multirow{2}{*}{\textit{Last-1}} \\ \cline{2-3}
 & SGF & MAS &  &  &  &  &  &  &  &  &  &  \\ \hline
(a) & \ding{55} & \ding{55} & 69.13\textcolor{red}{\tiny   \textdownarrow14.38} & 9.06\textcolor{red}{\tiny   \textdownarrow47.99} & 2.61\textcolor{red}{\tiny \textdownarrow73.45} & 84.60\textcolor{red}{\tiny   \textdownarrow2.02} & 63.18\textcolor{red}{\tiny \textdownarrow21.07} & 12.04\textcolor{red}{\tiny \textdownarrow70.52} & 84.93\textcolor{red}{\tiny   \textdownarrow1.91} & 46.51\textcolor{red}{\tiny   \textdownarrow33.04} & 84.93\textcolor{red}{\tiny   \textdownarrow1.91} & 2.61\textcolor{red}{\tiny   \textdownarrow54.44} \\
(b) & \ding{51} & \ding{55} & 70.50\textcolor{red}{\tiny \textdownarrow13.01} & 14.24\textcolor{red}{\tiny \textdownarrow42.81} & 7.01\textcolor{red}{\tiny \textdownarrow69.05} & 86.34\textcolor{red}{\tiny \textdownarrow0.28} & 65.03\textcolor{red}{\tiny \textdownarrow19.22} & 14.29\textcolor{red}{\tiny \textdownarrow68.27} & 86.50\textcolor{red}{\tiny \textdownarrow0.34} & 49.13\textcolor{red}{\tiny \textdownarrow30.42} & 86.50\textcolor{red}{\tiny \textdownarrow0.34} & 7.01\textcolor{red}{\tiny \textdownarrow50.04} \\
(c) & \ding{51} & \ding{51} & 83.51 & 57.05 & 76.06 & 86.62 & 84.25 & 82.56 & 86.84 & 79.55 & 86.84 & 57.05 \\ \hline
\end{tabular}
		}
		\label{tab: ablation_studies_miou}
	\end{table*}
	
	\begin{table*}[htb]
		\centering
		\caption{F1-score (\%) on the \ISPRS{} for Ablation Studies of SGF and MAS Modules. Each \textbf{Variant} corresponds to a different combination of components. \textit{Average}, \textit{Top-1}, and \textit{Last-1} indicate the average, highest, and lowest mIoU among all arbitrary-modal combinations, respectively. The modalities are denoted as follows: RGB (R), DSM (D), and SAR (S). The red arrows and values indicate the performance drop compared to \textbf{Variant (c)}, while the green arrows denote the improvement.}
		\resizebox{\linewidth}{!}{
			\begin{tabular}{c|cc|cccccccccc}
\hline
\multirow{2}{*}{Variants} & \multicolumn{2}{c|}{Components} & \multirow{2}{*}{R} & \multirow{2}{*}{D} & \multirow{2}{*}{N} & \multirow{2}{*}{RD} & \multirow{2}{*}{RN} & \multirow{2}{*}{DN} & \multirow{2}{*}{RDN} & \multirow{2}{*}{\textit{Average}} & \multirow{2}{*}{\textit{Top-1}} & \multirow{2}{*}{\textit{Last-1}} \\ \cline{2-3}
 & SGF & MAS &  &  &  &  &  &  &  &  &  &  \\ \hline
(a) & \ding{55} & \ding{55} & 81.11\textcolor{red}{\tiny   \textdownarrow9.81} & 14.68\textcolor{red}{\tiny   \textdownarrow55.68} & 5.03\textcolor{red}{\tiny \textdownarrow81.21} & 91.51\textcolor{red}{\tiny   \textdownarrow1.37} & 76.23\textcolor{red}{\tiny \textdownarrow15.11} & 18.83\textcolor{red}{\tiny \textdownarrow71.35} & 91.70\textcolor{red}{\tiny   \textdownarrow1.26} & 54.16\textcolor{red}{\tiny   \textdownarrow33.68} & 91.70\textcolor{red}{\tiny   \textdownarrow1.26} & 5.03\textcolor{red}{\tiny   \textdownarrow65.33} \\
(b) & \ding{51} & \ding{55} & 82.24\textcolor{red}{\tiny \textdownarrow8.68} & 22.28\textcolor{red}{\tiny \textdownarrow48.08} & 12.69\textcolor{red}{\tiny \textdownarrow73.55} & 92.51\textcolor{red}{\tiny \textdownarrow0.37} & 77.89\textcolor{red}{\tiny \textdownarrow13.45} & 24.74\textcolor{red}{\tiny \textdownarrow65.44} & 92.61\textcolor{red}{\tiny \textdownarrow0.35} & 57.85\textcolor{red}{\tiny \textdownarrow29.99} & 92.61\textcolor{red}{\tiny \textdownarrow0.35} & 12.69\textcolor{red}{\tiny \textdownarrow57.67} \\
(c) & \ding{51} & \ding{51} & 90.92 & 70.36 & 86.24 & 92.88 & 91.34 & 90.18 & 92.96 & 87.84 & 92.96 & 70.36 \\ \hline
\end{tabular}
		}
		\label{tab: ablation_studies_f1}
	\end{table*}
	
	\subsubsection{Progressive Component Analysis}
	
	The progressive integration of components reveals their distinct functional roles and demonstrates the necessity of both SGF and MAS modules for effective IMSS. The quantitative results are presented in \cref{tab: ablation_studies_miou} and \cref{tab: ablation_studies_f1}, showing mIoU and F1-score metrics for the \textbf{ISPRS} dataset 
	\begin{itemize}
		\item \textbf{Variant (a)} serves as the baseline direct additive fusion, suffering from severe modality imbalance as evidenced by extremely poor \textit{Last-1} performance. This validates the fundamental challenge where robust modalities dominate training while fragile modalities remain underutilized.
		\item \textbf{Variant (b)} introduces SGF without MAS, showing improvements across all modality configurations but with minimal gains in average metrics. While SGF enhances semantic consistency through global prototypes and adaptive fusion, the lack of targeted fragile-modal training limits overall effectiveness, particularly evident in the modest \textit{Average} improvement despite consistent gains across individual configurations.
		\item \textbf{Variant (c)} introduces both SGF and MAS, which achieves dramatic improvements in \textit{Average} mIoU and \textit{Last-1} mIoU. This superiority stems from MAS's robustness-guided sampling mechanism, which increases the training frequency of fragile modalities, thereby preventing information-rich, robust modalities from overshadowing information-sparse, fragile modalities, as occurs in conventional fusion.
	\end{itemize}
	
	Overall, the progression demonstrates that SGF fundamentally transforms multimodal interaction from fixed-weight fusion to adaptive semantic-guided integration. At the same time, MAS addresses the critical modality imbalance problem through targeted sampling strategies.

	\subsubsection{Abilities of SGF in Reducing Intra-class Variance}
	
	\begin{figure}[htb]
		\centering
		\includegraphics[width=\linewidth]{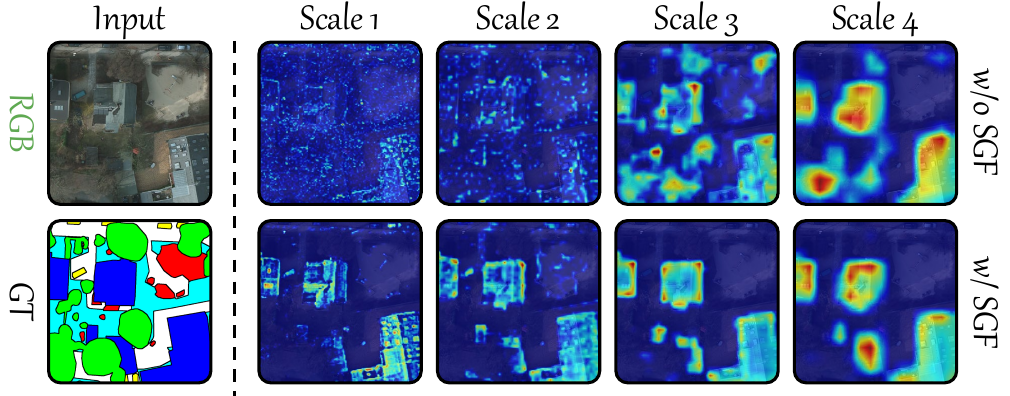}
		\caption{Grad-CAM based visualization of semantic activations on building-relevant regions in \textbf{Variant (a)} (w/o SGF) and \textbf{Variant (b)} (w/ SGF).}
		\label{fig: cam_visualize}
	\end{figure}
	
	\begin{figure}[htb]
		\centering
		\includegraphics[width=0.8\linewidth]{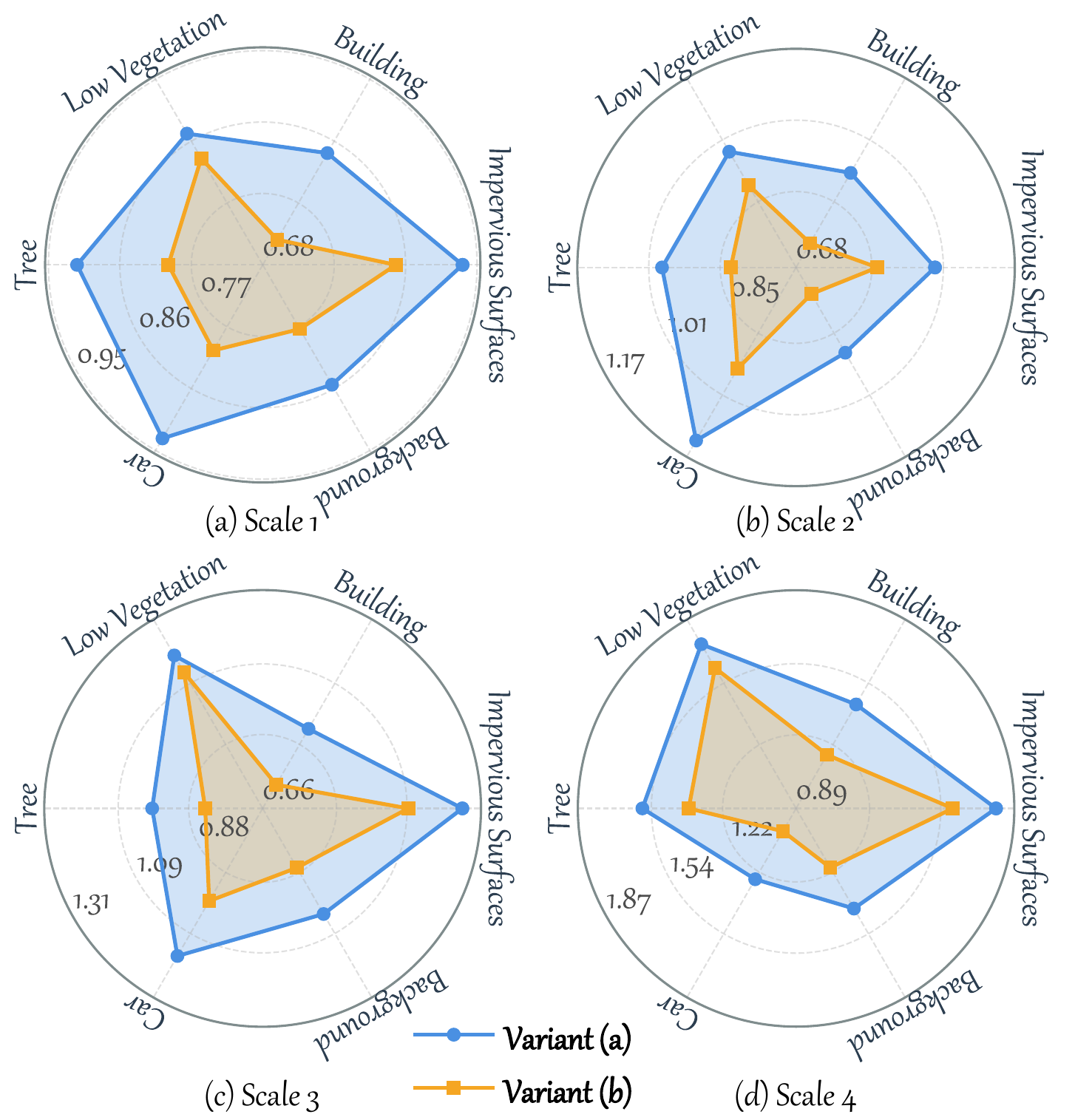}
		\caption{Intra-class variance comparison across different scales in \textbf{Variant (a)} (w/o SGF) and \textbf{Variant (b)} (w/ SGF).}
		\label{fig: variance_plots}
	\end{figure}

	To validate SGF's capability in reducing intra-class variance, we visualize features by Grad-CAM and compute quantitative variance metrics across \textbf{Variant (a)} without SGF and \textbf{Variant (b)} with SGF. As shown in \cref{fig: cam_visualize}, SGF enables significantly improved focus on target buildings regardless of their individual sizes and orientations across all scales. At Scale 1 and Scale 2, scattered activations in irrelevant background regions are substantially reduced. At Scale 3 and Scale 4, the semantic-guided fusion features exhibit markedly concentrated responses over entire building structures, achieving more precise boundaries and unified semantic representations. Quantitatively, the radar charts in \cref{fig: variance_plots} reveal that \textbf{Variant (b)} consistently achieves lower intra-class variance compared to \textbf{Variant (a)} across all four scales. Classes with diverse appearances show the most significant reduction: \textit{building} variance decreases from 0.84 to 0.74 at Scale 1 and from 1.43 to 1.17 at Scale 4, while \textit{car} variance reduces from 1.13 to 0.94 at Scale 1 and from 1.26 to 1.01 at Scale 4. More homogeneous classes such as \textit{low vegetation} exhibit modest yet consistent improvements. This demonstrates that SGF's semantic-guided perception mechanism leverages global semantic prototypes as class-specific anchors, effectively harmonizing diverse appearances into consistent semantic representations while maintaining discriminative inter-class boundaries.
	
	\subsubsection{Abilities of SGF in Harmonizing Cross-modal Heterogeneity}
	
	\begin{figure}[htb]
		\centering
		\includegraphics[width=\linewidth]{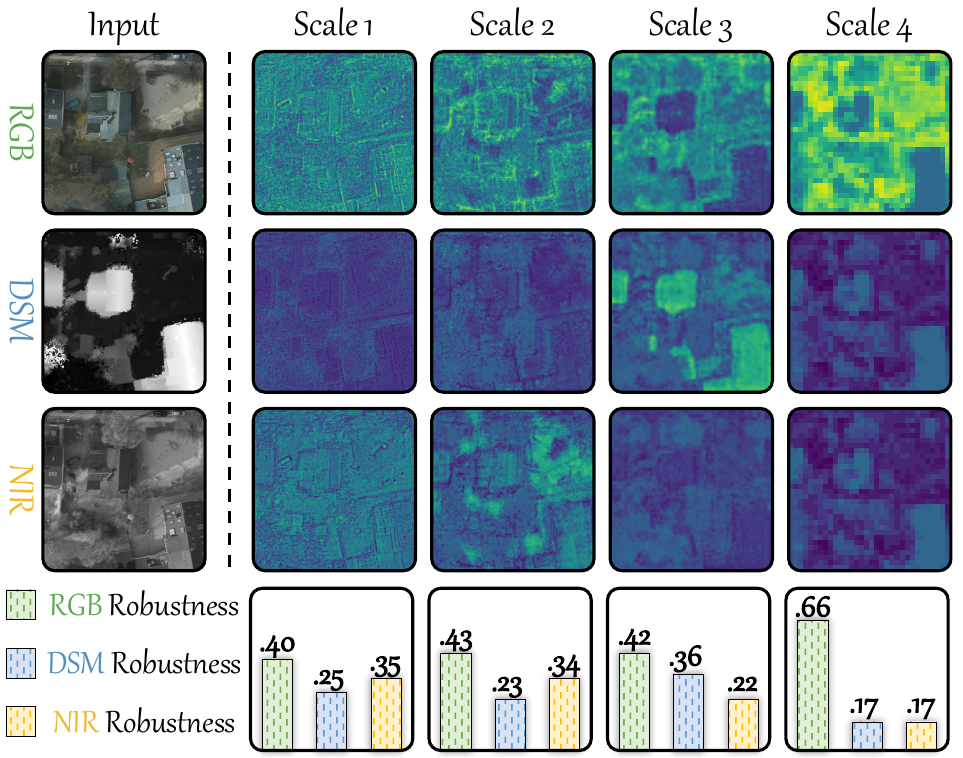}
		\caption{Visualization of robustness maps and average scalars across all scales.}
		\label{fig: robustness_visualize}
	\end{figure}
	
	To investigate SGF's ability to mitigate cross-modal heterogeneity through adaptive modality weighting, we visualize the robustness maps and corresponding average scalars across all scales in \cref{fig: robustness_visualize}. The robustness maps demonstrate SGF's scale-adaptive perception of modality reliability. At Scale 1 and Scale 2, RGB maintains moderate dominance at 0.40-0.43 while DSM and NIR contribute balanced weights, indicating that fine-grained spatial details benefit from multi-modal complementarity. At Scale 3, DSM's contribution of 0.36 nearly matches RGB's 0.42, suggesting structural height information becomes critically important for mid-level semantic understanding. At Scale 4, RGB dramatically dominates at 0.66 while other modalities contribute minimally at 0.17 each, reflecting the primacy of high-level semantic context for final classification. This dynamic modality weighting mechanism directly addresses cross-modal heterogeneity by intelligently emphasizing the most reliable modality at each scale while preserving complementary information from others, enabling SGF to harmonize cross-modal inconsistencies while maintaining each modality's unique semantic contributions.
	
	\subsubsection{Abilities of MAS in Mitigating Multimodal Imbalance}
	
	\begin{figure}[htb]
		\centering
		\includegraphics[width=\linewidth]{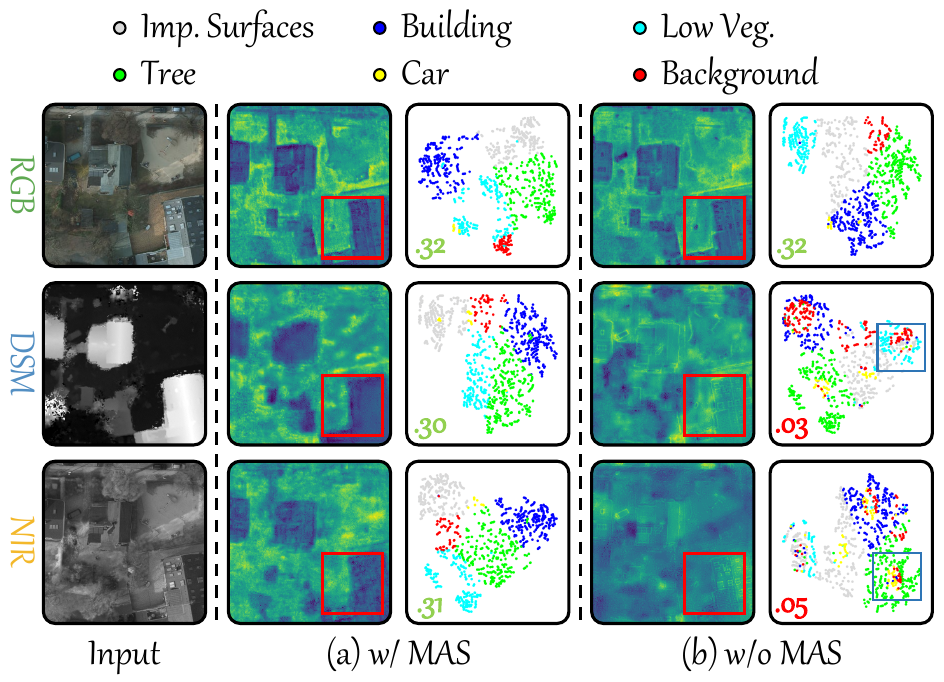}
		\caption{Comparison of feature maps and t-SNE embeddings under the conditions (a) with MAS and (b) without MAS. The t-SNE visualizations project high-dimensional features into a 2D space, where different colors represent distinct semantic categories, and the bottom-left values denote silhouette scores, which measure cluster separability. Compact clusters with higher silhouette scores indicate better feature discriminability, while entangled distributions with lower scores suggest class confusion.}
		
		\label{fig: feature_visualize}
	\end{figure}
	
	\begin{table}[htb]
		\centering
		\caption{Computational and parameter complexity of different methods, where backbone is excluded from the statistics.}
		\label{tab: efficiency_comparison}
		\begin{tabular}{l|cc}
	\hline
	Methods & \#FLOPs (G) & \#Params (M) \\ \hline
	\MuSS{} & 27.00 & 58.58 \\
	\MtL{} & \textbf{8.68} & \textbf{3.55} \\
	\IMLT{} & 10.44 & {\ul 4.33} \\
	\MAGIC{} & 98.11 & 22.29 \\
	SGMA (\textbf{Ours}) & {\ul 9.47} & 4.79 \\
	\rowcolor[HTML]{D9D9D9} 
	\wrtSOTA{} & +0.79 & +1.24 \\ \hline
\end{tabular}
	\end{table}
	
	\begin{figure}[htb]
		\centering
		\includegraphics[width=\linewidth]{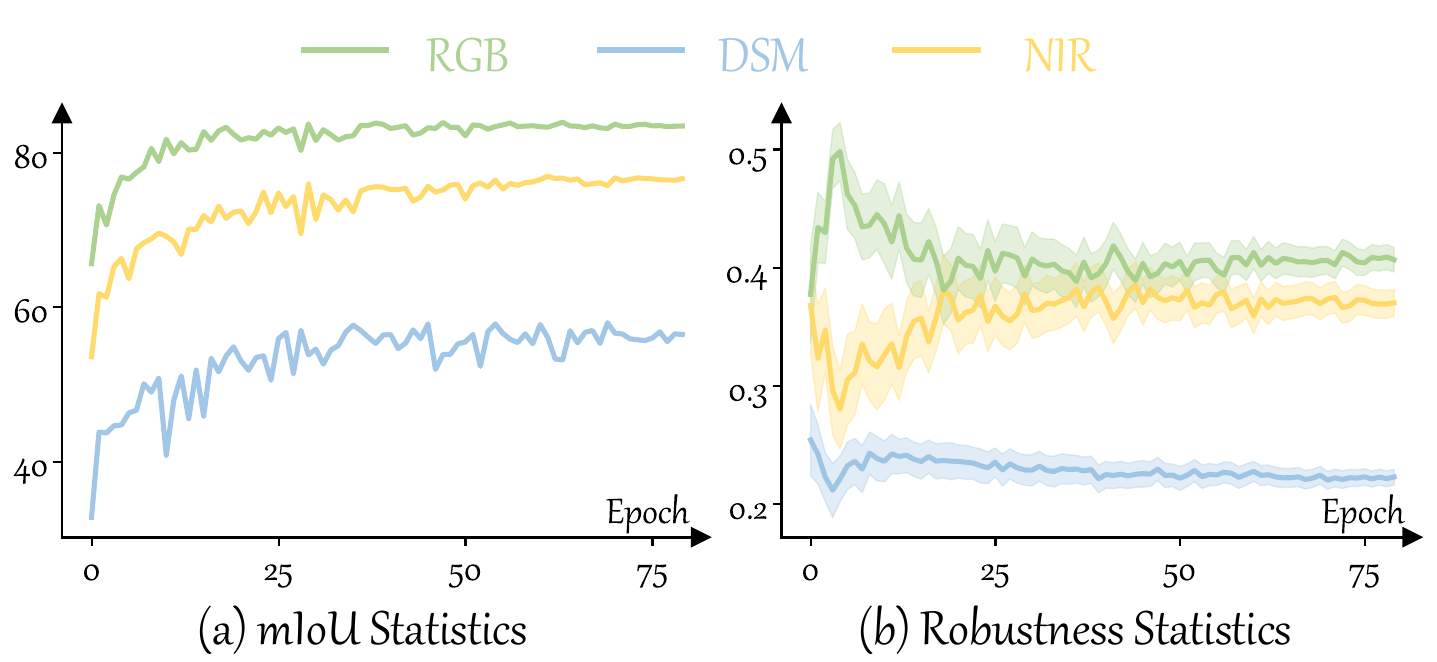}
		\caption{Training dynamics analysis on ISPRS dataset with PVT-v2-b2 backbone at scale level 2.}
		\label{fig: robustness_epoch_visualize}
	\end{figure}
	
	\begin{figure*}[htb]
		\centering
		\includegraphics[width=0.8\linewidth]{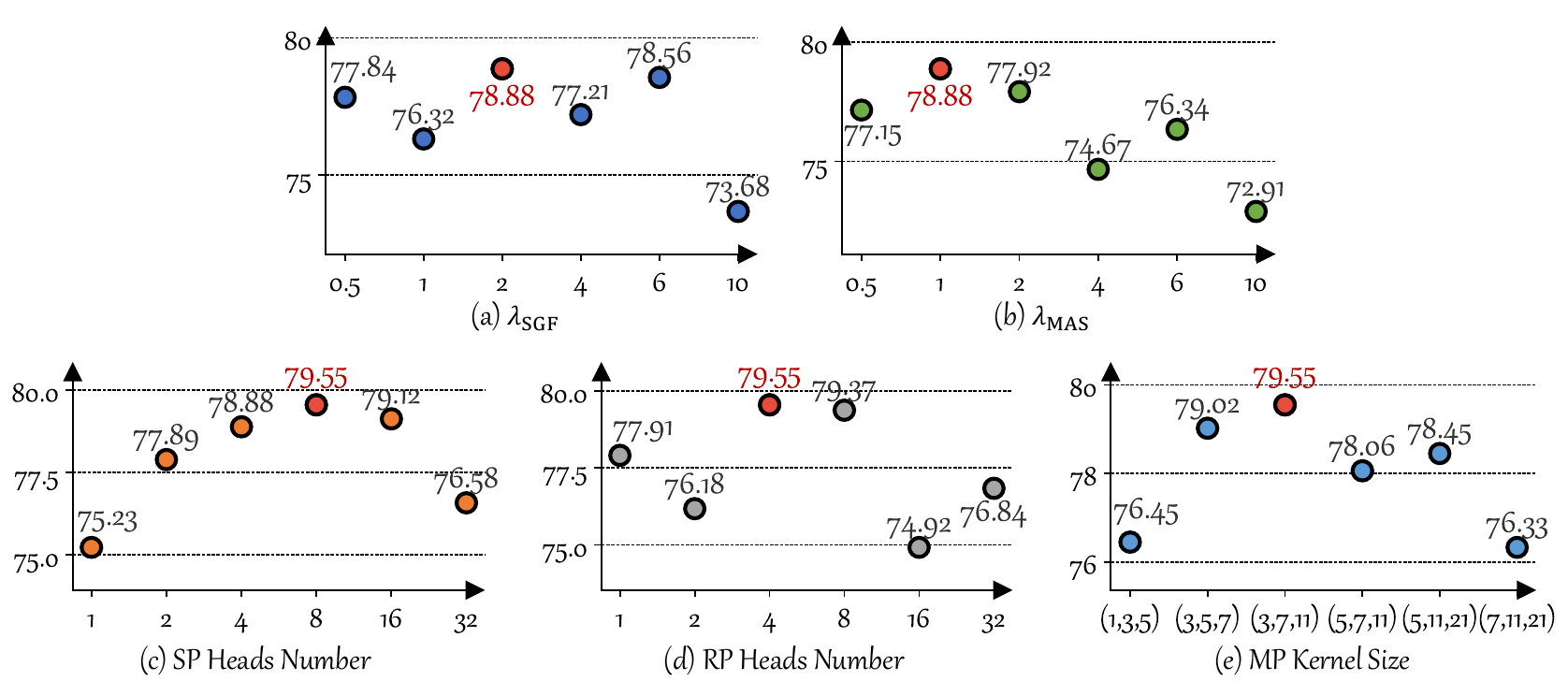}
		\caption{Component-wise ablation analysis of SGF and MAS. Sequential optimization of (a) $\lambda_\text{SGF}$, (b) $\lambda_\text{MAS}$, (c) SP heads, (d) RP heads, and (e) MP kernel sizes. Red circles denote optimal values that maximize mIoU (\%).}
		\label{fig: parameter_ablation_studies}
	\end{figure*}
	
	To evaluate MAS's contribution in mitigating multimodal imbalance, we conduct feature visualization analysis comparing \textbf{Variant (b)} with SGF only and \textbf{Variant (c)} with both SGF and MAS. As illustrated in \cref{fig: feature_visualize}, \textbf{Variant (b)} without MAS exhibits weak and spatially diffuse responses in DSM and NIR modalities, particularly around building boundaries (red boxes), with severely entangled t-SNE embeddings (blue boxes) where different semantic classes are heavily overlapped. This severe class confusion is quantitatively reflected in silhouette scores, where only the robust RGB modality achieves reasonable cluster separability with a score of 0.32, while the fragile DSM and NIR modalities obtain extremely low scores of 0.03 and 0.05, respectively, indicating near-random cluster assignments and insufficient discriminative learning. In contrast, \textbf{Variant (c)} with MAS demonstrates substantial improvements where multimodal semantic feature maps become spatially coherent with clearer activations, and t-SNE embeddings exhibit dramatically improved structure with compact intra-class clusters and distinct inter-class separation. Quantitatively, MAS enables all three modalities to achieve consistently high silhouette scores of 0.32, 0.30, and 0.31 respectively, demonstrating that fragile modalities now learn discriminative representations without compromising the robust modality. This results from MAS's robustness-guided sampling mechanism, which leverages SGF's reliability estimates to increase the sampling frequency of fragile modalities, ensuring they receive adequate training attention and develop unique discriminative capabilities rather than being overshadowed by robust modalities, ultimately addressing the core problem of multimodal imbalance.
	
	\subsubsection{Complexity Analysis}
	
	To evaluate the scalability of SGMA for practical applications, we analyze the computational and parameter complexity. As shown in \cref{tab: efficiency_comparison}, SGMA achieves competitive efficiency with only 0.79 FLOPs and 1.24 million parameters, surpassing the most lightweight methods. Compared to the PVT-v2-b2~\cite{wang2022pvt} backbone requiring 72.92 GFLOPs and 24.85 million parameters, this modest overhead accounts for increases of only 1.1\% and 1.7\%, respectively. Notably, both computational cost and parameters scale linearly with spatial dimensions ($H_i$, $W_i$), number of classes ($K$), and number of modalities ($M$), demonstrating excellent scalability of our framework. This confirms that SGMA maintains a favorable balance between performance and computational efficiency, making it suitable for practical multimodal segmentation applications.
	
	\subsubsection{Training Dynamics and Robustness Score Analysis}
	
	To comprehensively validate our framework's effectiveness, we analyze the training dynamics by tracking both segmentation performance and robustness scores at scale level 2 across 80 epochs. As shown in \cref{fig: robustness_epoch_visualize}, three key observations emerge:
	
	\begin{enumerate}
		\item \textbf{Robustness-Performance Correlation:} The ranking of robustness scores in (b) (RGB > NIR > DSM) is highly consistent with the ranking of mIoU in (a) throughout training, validating that our robustness metric effectively captures modality reliability and informativeness.
		
		\item \textbf{Balanced Convergence:} All modalities converge within 30 epochs with satisfactory performance. Robust modalities (RGB, NIR) stabilize earlier around epoch 10-15, while the fragile modality (DSM) shows continuous improvement without premature saturation. This demonstrates that MAS and SGF's complementary design ensures balanced learning: MAS increases training frequency for fragile modalities while SGF adaptively emphasizes robust modalities when reliable, enabling all modalities to be sufficiently exploited without sacrificing convergence speed or accuracy.
		
		\item \textbf{Stability Evolution:} The shaded regions in (b) show decreasing standard deviations across samples as training progresses. By epoch 30, all modalities reach stable robustness values with minimal fluctuation, synchronized with segmentation performance convergence. This confirms that our robustness metric reliably reflects modality learning dynamics and provides consistent guidance for both SGF and MAS throughout training.
	\end{enumerate}
	
	\subsubsection{Hyperparameter Sensitivity Analysis}
	
	To find the optimal hyperparameters, we conduct extensive ablation studies on configurable hyperparameters. Starting from default settings of $\lambda_\text{SGF}$=1, $\lambda_\text{MAS}$=1, SP Heads=4, RP Heads=4, and MP Kernel Size=(3,7,11), we adopt a sequential search strategy as shown in \cref{fig: parameter_ablation_studies}. We first optimize $\lambda_\text{SGF}$ and find that setting it to 2 achieves 78.88\% mIoU. We then optimize $\lambda_\text{MAS}$ and find that keeping it at 1 maintains this performance. Next, we sequentially optimize the number of SP heads and RP heads, where 8 SP heads and 4 RP heads both reach 79.55\% mIoU. Finally, the MP kernel size of (3, 7, 11) confirms the optimal configuration at 79.55\% mIoU.
	
	\section{Conclusion} \label{sec: conclusion}
	
	This paper addresses three fundamental challenges in IMSS for remote sensing: modality imbalance where robust modalities dominate training, substantial intra-class variation across diverse spatial scales and orientations, and firm cross-modal heterogeneity leading to conflicting semantic cues. We propose SGMA with two synergistic components that enable robust segmentation under arbitrary modality-missing scenarios. SGF addresses intra-class variation by extracting global semantic prototypes as class-specific anchors, harmonizing diverse appearances while providing reliable modality robustness estimates and reconciling cross-modal heterogeneity through adaptive attention-weighted fusion. MAS leverages these robustness cues to dynamically increase training frequencies of fragile modalities, enhancing their discriminative capabilities while preventing dominance by robust modalities. Extensive experiments on \ISPRS{}, \DFC{}, and \DELIVER{} confirm that our framework successfully exploits complementary multimodal information while maintaining robustness across diverse sensor configurations, providing a principled solution for IMSS in remote sensing applications with both theoretical insights and practical benefits for real-world deployment where sensor reliability varies. Future work will address the current lack of interpretability by developing mechanisms to explicitly quantify modality-specific learning dynamics and explore extensions to broader remote sensing scenarios and temporal multimodal sequences.
	
	\bibliographystyle{IEEEtran}
	\bibliography{ref}

    \begin{IEEEbiography}[{\includegraphics[width=1in,height=1.25in,clip,keepaspectratio]{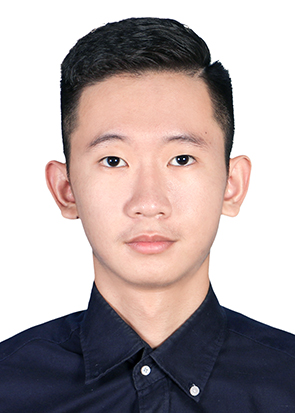}}]{Lekang Wen} received the B.S. degree from Sun Yat-sen University, Zhuhai, China, in 2024. He is currently pursuing the Ph.D. degree with the State Key Laboratory of Information Engineering in Surveying, Mapping and Remote Sensing, Wuhan University, Wuhan, China. His research interests include multimodal learning, foundation models, and satellite photogrammetry.
	\end{IEEEbiography}
    
	\begin{IEEEbiography}[{\includegraphics[width=1in,height=1.25in,clip,keepaspectratio]{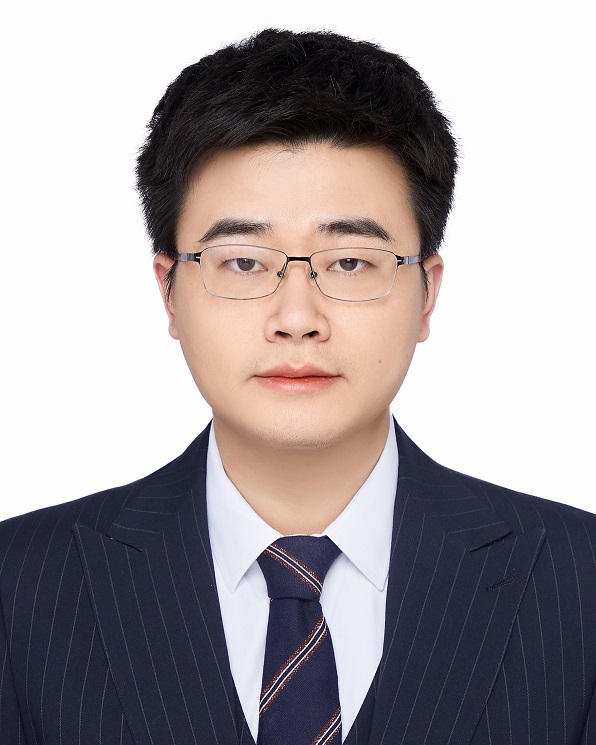}}]{Liang Liao} (Senior Member, IEEE) received the B.S. degree from International School of Software, Wuhan University, Wuhan, China, in 2013, and the Ph.D. degree from the National Engineering Research Center for Multimedia Software, School of Computer Science, Wuhan University, Wuhan, China, in 2019. He is currently a Professor with the Hangzhou Institute of Technology, Xidian University. He was a Research Fellow with the School of Computer Science and Engineering, Nanyang Technological University, Singapore, from 2022 to 2024, and was a Project Researcher with the National Institute of Informatics, Japan, from 2019 to 2022. His research interests include image/video processing, transmission, and quality assessment.
	\end{IEEEbiography}

    \begin{IEEEbiography}[{\includegraphics[width=1in,height=1.25in,clip,keepaspectratio]{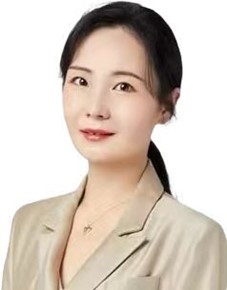}}]{Jing Xiao} received the B.Sc. and M.Sc. degrees in computer science and technology from Wuhan University, Wuhan, China, in 2006 and 2008, respectively, and the Ph.D. degree in computer science and technology from the Institute of Geo-Information Science and Earth Observation, Twente University, Enschede, The Netherlands, in 2013. She was a Project Researcher with the National Institute of Informatics, Tokyo, Japan, from 2019 to 2020. She is currently a Professor with the National Engineering Research Center for Multimedia Software, School of Artificial Intelligence, Wuhan University. Her research interests include image/video processing and compression and analysis.
	\end{IEEEbiography}

    \begin{IEEEbiography}[{\includegraphics[width=1in,height=1.25in,clip,keepaspectratio]{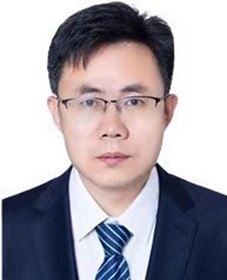}}]{Mi Wang} received the B.Eng., M.Sc., and Ph.D. degrees in photogrammetry and remote sensing from Wuhan University, Wuhan, China, in 1997, 1999, and 2001, respectively. Since 2008, he has been a Professor with the State Key Laboratory of Information Engineering in Surveying, Mapping and Remote Sensing, Wuhan University. His research interests include measurable seamless stereo ortho-image databases, geographic information systems (GIS), and high-precision remote-sensing image processing and intelligent spatial information service.
	\end{IEEEbiography}
	
\end{document}